\documentclass[10pt,twocolumn,letterpaper]{article}

\usepackage[pagenumbers]{format_cvpr2024/cvpr} % To force page numbers, e.g. for an arXiv version

\usepackage[accsupp]{axessibility} % Improves PDF readability for those with disabilities.

\usepackage[dvipsnames]{xcolor}

\usepackage{multirow}
\usepackage{pifont}
\usepackage{amsfonts}
\usepackage{bm}

\newcommand{\R}{\mathbb{R}}

\makeatletter
\def\thickhline{%
  \noalign{\ifnum0=`}\fi\hrule \@height \thickarrayrulewidth \futurelet
   \reserved@a\@xthickhline}
\def\@xthickhline{\ifx\reserved@a\thickhline
               \vskip\doublerulesep
               \vskip-\thickarrayrulewidth
             \fi
      \ifnum0=`{\fi}}
\makeatother

\newlength{\thickarrayrulewidth}
\usepackage[toc,page,header]{appendix}
\usepackage{minitoc}

\definecolor{cvprblue}{rgb}{0.21,0.49,0.74}
\usepackage[pagebackref,breaklinks,colorlinks,citecolor=cvprblue]{hyperref}

\def\paperID{2549} % *** Enter the Paper ID here
\def\confName{CVPR}
\def\confYear{2024}

\title{UnSAMFlow: Unsupervised Optical Flow Guided by Segment Anything Model}

\author{Shuai Yuan\thanks{Affiliated with Meta at the time of this work.}~, Lei Luo, Zhuo Hui, Can Pu, Xiaoyu Xiang, Rakesh Ranjan, and Denis Demandolx\\
Meta Reality Labs\\
{\tt\small \{shuaiyuan,luoleyouluole,harryhui,cpu,xiangxiaoyu,rakeshr,denisd\}@meta.com}
}

\begin{document}

% PAGE LIMIT: 8
\maketitle

\begin{abstract}
Traditional unsupervised optical flow methods are vulnerable to occlusions and motion boundaries due to lack of object-level information. Therefore, we propose UnSAMFlow, an unsupervised flow network that also leverages object information from the latest foundation model Segment Anything Model (SAM). We first include a self-supervised semantic augmentation module tailored to SAM masks. We also analyze the poor gradient landscapes of traditional smoothness losses and propose a new smoothness definition based on homography instead. A simple yet effective mask feature module has also been added to further aggregate features on the object level. With all these adaptations, our method produces clear optical flow estimation with sharp boundaries around objects, which outperforms state-of-the-art methods on both KITTI and Sintel datasets. Our method also generalizes well across domains and runs very efficiently. %Code is available at \url{https://github.com/facebookresearch/UnSAMFlow}.
\end{abstract}    
\section{Introduction}

% 1. optical flow is important; many applications.
% \HH{A number of applications, including autonomous driving~\cite{geiger2012we,yuan2023semarflow}, video processing~\cite{} and image editing~\cite{}, require the per-pixel correspondence. This is classically done by solving the optical flow problem, typically in the form of per-pixel shift in horizontal and vertical directions. Because of the ill-posed nature of this problem, optical flow methods rely on additional prior ~\cite{lukas-kanade, tv-l1....} and are prone to errors in textureless and non-Lambertian regions.} 

Optical flow estimation~\cite{lucas1981iterative,horn1981determining} involves finding pixel-level correspondences between video frames, which has broad applications such as video understanding~\cite{ye2022deformable}, video editing~\cite{gao2020flow,kim2022cross}, and
autonomous driving~\cite{geiger2012we,yuan2023semarflow}.

% 2. a lot of progress on supervised; label is expensive $\to$ unsup optical flow
% \HH{More recently, researchers have looked at the problem under relatively unconstrained conditions, and using dense annotated data with supervised learning. Several optical flow algorithms have been developed along this line~\cite{dosovitskiy2015flownet,sun2018pwc,hur2019iterative,teed2020raft,zhang2021separable,jiang2021learning,luo2022learning,jung2023anyflow,huang2022flowformer,shi2023flowformer++}. The drawback of these approaches is that the ground-truth label is difficult to acquire and requires precisely calibrated, and often prohibitively expensive, acquisition setups. This makes these techniques, which are trained specific to each task, difficult to apply at large-scales.}
Following the latest trend of deep learning in computer vision~\cite{krizhevsky2012imagenet,he2016identity,dosovitskiy2020image}, most recent methods have modeled the optical flow problem under the supervised learning framework~\cite{dosovitskiy2015flownet,sun2018pwc,hur2019iterative,teed2020raft,zhang2021separable,jiang2021learning,luo2022learning,jung2023anyflow,huang2022flowformer,shi2023flowformer++}, where ground-truth labels are used to train the networks. However, obtaining such labels for real-life videos is especially difficult since it usually requires precise calibrations across multiple sensors, leading to prohibitively high annotation costs~\cite{yuan2022optical}. This drawback makes these supervised techniques hard to be applied to large-scale real applications.

% 3. unsup optical flow is done by photometric, issues with occlusions, so add smoothness loss to regularize the occluded flow; smoothness does not work at boundaries
% \HH{One solution to this problem is to explicitly account brightness constancy with smoothness prior~\cite{yu2016back,ren2017unsupervised,meister2018unflow,liu2020learning,jonschkowski2020matters,luo2021upflow,stone2021smurf}. While effective, these techniques are inherently limited by the assumption on the occlusion regions, where the motion is cut off abruptly. Recent attempts~\cite{hur2016joint,wulff2017optical,sevilla2016optical,yuan2023semarflow} at addressing occlusion issue use semantic segmentation to resolve the ambiguity across the object boundaries. However, these approaches do not distinguish instances within the same semantic class, which greatly limits the capability in resolving occlusion.}

% Different from supervised training, unsupervised flow networks rely on the brightness constancy and smoothness of the flow field to define losses~\cite{yu2016back,ren2017unsupervised,meister2018unflow,liu2020learning,jonschkowski2020matters,luo2021upflow,stone2021smurf}. Specifically, the corresponding points across frames should have similar local appearances, and the optical flow field should be smooth. 
Due to the high annotation costs, much recent work has focused on the \emph{unsupervised} training of optical flow~\cite{yu2016back}.
Instead of ground-truth labels, unsupervised flow networks rely on two key principles to define losses~\cite{yu2016back,ren2017unsupervised,meister2018unflow,liu2020learning,jonschkowski2020matters,luo2021upflow,stone2021smurf}. Firstly, brightness constancy assumes that the corresponding points across frames should maintain similar local appearances. Secondly, the optical flow field should be spatially smooth. However, these assumptions are compromised at occlusion regions~\cite{wang2018occlusion,janai2018unsupervised}, where foreground objects cover background appearances, and around motion boundaries~\cite{yu2022unsupervised,yang2022decomposing}, where the motion is cut off abruptly. These issues are pervasive in real applications and have posted great challenges to unsupervised optical flow~\cite{yuan2023assisting}.

\begin{figure}
    \centering
    \includegraphics[width=\linewidth]{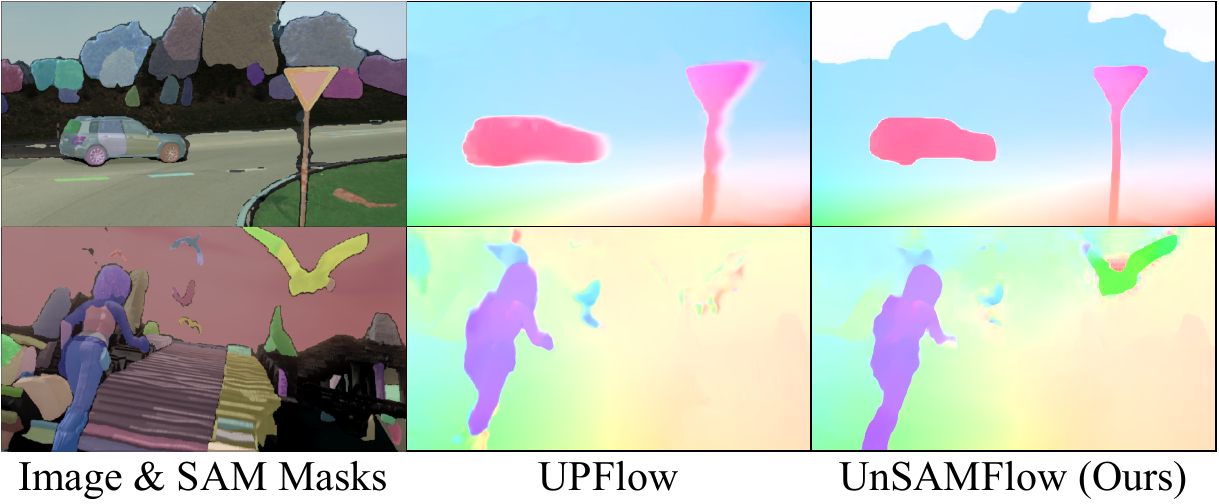}
    \caption{Our UnSAMFlow utilizes object-level information from SAM~\cite{sam} to generate clear optical flow with sharp boundaries.} 
    \label{fig:intro}
\end{figure}
% \HH {It would be better to use tabular which makes the text look better}
% \setlength{\abovecaptionskip}{10pt}
% \setlength{\belowcaptionskip}{0pt}

% 4. optical flow is a low-level, pixel-level task; can we add higher-level object information to help? object information helps understand occlusions; object info helps better define smoothness (piecewise); help extract features on the object level 
Fundamentally, the issues with occlusions and motion boundaries both stem from the \emph{low-level} nature of optical flow, where \emph{object-level} information is generally missing. To better handle occlusions, it is important to understand the spatial relationships and interactions between objects. Also, optical flow should be smooth only within the same continuous object region, while sharp motion boundaries are allowed near object edges. Thus, object-level information could play a key role in refining unsupervised optical flow.

% 5. early works add semantics to help, but mostly semantic segmentation, causes issues because different instances of objects are not separated. different instances can have totally different motions $\to$ more instance based object masks,

Indeed, some previous methods have explored aggregating object information, using semantic segmentation to help optical flow~\cite{hur2016joint,wulff2017optical,sevilla2016optical,yuan2023semarflow}. However, though convenient, the use of semantic segmentation is not precise because it does not distinguish different instances of the same semantic class, which may have drastically different motions. It is also constrained by the limited number of classes defined and may not recognize novel objects in the open world.

% 6. strength of SAM as semantic information provider; moti ation of using SAM differentiate from previous work, SAM provides instance information (not instance segmentation because no semantic class) and is general-purpose, good zero-shot performance, can work across domains; SAM works on all object levels, e.g. separating body parts of each person. $\to$ use SAM output to help unsup flow 
In comparison, the latest Segment Anything Model~\cite{sam} (SAM) may be a better option. SAM is a general-purpose image segmentation model pre-trained on very large and diverse datasets. It can separate different instances and has shown impressive zero-shot performances on objects not seen in training. In addition, SAM detects objects of various scales and levels, segmenting small object parts (such as hands and arms) as well. This can reduce complexity and help differentiate motions of object parts separately. 

So motivated, we integrate SAM as additional object-level information to enhance unsupervised optical flow, which can be achieved through three novel adaptations. We first adapt the semantic augmentation module from SemARFlow~\cite{yuan2023semarflow} to enable self-supervision based on SAM masks (\cref{subsec:aug}). Moreover, we enforce smooth motion within each SAM segment using a new regional smoothness loss based on homography (\cref{subsec:hg}). This approach effectively rectifies numerous inconsistent flow outliers. Lastly, we design a mask feature module to aggregate features from the same SAM mask for robustness (\cref{subsec:mf}).

% 9. results
Our method significantly outperforms previous methods on both KITTI~\cite{kitti12,kitti15} and Sintel~\cite{sintel} benchmarks through both quantitative (\cref{subsec:benchmark}) and qualitative (\cref{subsec:qual}) evaluations. Notably, our network achieves 7.83\% test error on KITTI-2015~\cite{kitti15}, outperforming the state-of-the-art UPFlow~\cite{luo2021upflow} (9.38\%) and SemARFlow~\cite{yuan2023semarflow} (8.38\%) by a clear margin. As the examples show in \cref{fig:intro}, our method produces much clearer and sharper motion that is consistent with the SAM masks. Extensive ablation studies also justify the effectiveness of each proposed adaptation (\cref{subsec:ablate}). Further analysis shows that our method generalizes well across domains (\cref{subsec:generalize}) and runs efficiently (\cref{subsec:time}).

% 10. summary of contribution: (1) the first method to incorportate SAM with unsup optical flow; (2) analyze the issue of widely used boundary-aware smoothness loss and propose our homography smoothness; (3) ways to process SAM masks that can be used in other tasks; (4) give code for future research.

% \LL{For contribution, should we also mention that our method achieves good performance by incorporating SAM masks, but it is only used for training, not for inference, so the inference time is still low while having good quality?}

% We should combine it with unsupervised with semantic 
% we should also motive the use of homography

% \begin{itemize}
%     \item We introduce SAM~\cite{sam} with unsupervised learning for optical flow estimation.
%     \item We present the new loss by pooling together information within instance, enabling robust estimation on object boundary and occlusion.
%     \item  We demonstrate strong performance on cross-dataset generalization by achieving more than 5\% of error rate reduction compared with state-of-the-art methods and rank 1st on both KITTI and Sintel benchmark for unsupervised learning.
% \end{itemize}

% \HH{
% Together, our contributions provide, for the first time, an approach for using SAM~\cite{sam}, which significantly improves the quality for unsupervised optical flow estimation and extends the applicability for wide range real world applications without ground truth labels.}

In summary, our contributions are as follows.
\begin{itemize}
    \item To the best of our knowledge, we are the first to effectively combine SAM~\cite{sam} with unsupervised optical flow estimation, which helps learning optical flow for wide-range real-world videos without ground-truth labels.
    \item We analyze the issues of previous smoothness losses with visualizations and propose a new smoothness loss definition based on homography and SAM as a solution.
    \item We show how SAM masks can be processed, represented, and aggregated into neural networks, which can be directly extended to other tasks using SAM. 
    % \item We release full reproducible code for follow-up research.
\end{itemize}

\section{Related work}

% \HH{
% \paragraph{Supervised optical flow.} Deep neural network-based techniques have been proposed for estimating optical flow from a pair of images. These methods characterize the inference as image to image transform. These approaches can provide  precise and high-quality flow estimates, but require the ground truth data, making their applicability in real world  inherently limited.}

% \HH{
% \paragraph{Unsupervised optical flow.} The seminal work of Lucas and Kanade~\cite{} demonstrated the optical flow problem can be solved with brightness constancy. Follow up work~\cite{} has demonstrated that the use of local smooth prior is able to mitigate the occlusion issue. More recently, CNN-based methods have also been proposed for estimating the deep prior to restrict the underlying problem, including occlusion-aware adjustments~\cite{wang2018occlusion,janai2018unsupervised,meister2018unflow,liu2021oiflow,yuan2023ufd}, iterative refinement~\cite{hur2019iterative}, learned upsampler~\cite{luo2022learning}, self-supervision~\cite{liu2019ddflow,liu2019selflow,liu2020learning,stone2021smurf}, dataset learning~\cite{sun2021autoflow,han2022realflow,huang2023self} and multi resolution correlation volume~\cite{stone2021smurf}. However, the deep prior from these approaches is scene independent and in many cases, the motion and objects can be arbitrarily different from training. In contrast, we propose the use of semantic information which directly maps the prior with each scene and demonstrate that it leads to a better generalizability and flow estimates.}

\paragraph{Unsupervised optical flow}
% Despite the substantial recent progress on supervised optical flow training~\cite{teed2020raft,liu2021asflow,xu2022gmflow,sui2022craft,huang2022flowformer,wu2023accflow,shi2023flowformer++,lu2023transflow,saxena2023surprising,luo2023gaflow,jung2023anyflow}, unsupervised optical flow has received less attention in recent years.
Traditional methods~\cite{lucas1981iterative,horn1981determining,zach2007duality} optimize optical flow based on brightness constancy and local smoothness.
These constraints have been transformed to photometric and smoothness losses in early unsupervised networks~\cite{yu2016back,ren2017unsupervised}. Since then, more specific modules have been proposed, including occlusion-aware adjustments~\cite{wang2018occlusion,janai2018unsupervised,meister2018unflow,liu2021oiflow,yuan2023ufd}, iterative refinement~\cite{hur2019iterative}, learned upsampler~\cite{luo2022learning}, self-supervision~\cite{liu2019ddflow,liu2019selflow,liu2020learning,stone2021smurf}, dataset learning~\cite{sun2021autoflow,han2022realflow,huang2023self}. The latest SMURF~\cite{stone2021smurf} based on RAFT~\cite{teed2020raft} has also achieved outstanding performances.
% Still, our network is adapted from ARFlow~\cite{liu2020learning} due to its simplicity.

\paragraph{Segment Anything Model (SAM)}
Segment Anything Model (SAM)~\cite{sam} is a recent general-purpose foundation model for image segmentation tasks. The model is trained on enormous high-quality annotated images and is designed to accept prompts (points, boxes, \etc) to retrieve object masks. Its trained model can transfer well zero-shot to new data distributions and thus has been applied to many vision tasks and applications such as object tracking~\cite{yang2023track,cheng2023segment}, video segmentation~\cite{rajivc2023segment,zhang2023uvosam}, neural rendering~\cite{cen2023segment,shen2023anything,chen2023interactive}, and medical imaging~\cite{mazurowski2023segment,ma2023segment,wu2023medical,he2023accuracy}

% \HH{ 
% \paragraph{Relationship to prior work.}
% There have been other methods~\cite{sevilla2016optical,wulff2017optical,bai2016exploiting,yuan2023semarflow,wang2019semflow} similar to our approach that seek to incorporate semantic or instance segmentation models for optical flow estimation. The flow estimates is combined with semantic model to provide geometry prior, in the form of homography transform~\cite{sevilla2016optical}, epipolar constraint~\cite{bai2016exploiting}, and camera poses~\cite{wulff2017optical,sevilla2016optical}.   
% This transforms semantic labels into instance correspondence with introduced motion model, however, this is a coarse approximation and the semantic model is not able to distinguish instance with the same class. In this work, we address these challenges by incorporating SAM~\cite{sam} into the training pipeline and leveraging the results to better inference the local object motion.}

% \HH{ 
% The work closest to ours is the concurrently proposed SAMFlow~\cite{zhou2023samflow} algorithm of Zhou et al. While they rely on SAM~\cite{sam} like us, we do not require the ground truth label of optical flow, as well as the features from SAM model. Unlike SAMFlow~\cite{zhou2023samflow}, our technique does not require annotations for the flow. Nor does  it rely on any assumptions on SAM~\cite{sam}. Instead, we can automatically estimate the flow with SAM-style (@shuai, check whether it is the right word to use) object masks. 
% }

\paragraph{Combining optical flow and object information}
To aggregate object-level information, many previous methods have combined semantic or instance segmentation models off-the-shelf as object cues to help refine optical flow~\cite{sevilla2016optical,wulff2017optical,bai2016exploiting,yuan2023semarflow,wang2019semflow}. Given semantics, most methods reason the rigid motions of objects and conduct refinement based on geometric techniques such as homography~\cite{sevilla2016optical}, epipolar geometry~\cite{bai2016exploiting}, and SfM~\cite{wulff2017optical,sevilla2016optical}. SemARFlow~\cite{yuan2023semarflow} is a latest neural network that incorporates semantics on the feature level and through self-supervision, which we follow closely. Besides, joint training has also been explored to benefit both optical flow and segmentation tasks based on semantic consistency and occlusion reasoning~\cite{hur2016joint,cheng2017segflow,ding2020every}.

One concurrent work, SAMFlow~\cite{zhou2023samflow}, also combines SAM with optical flow. However, we focus on \emph{unsupervised} flow as opposed to their \emph{supervised} flow. Furthermore, our method imports SAM \emph{outputs} instead of SAM \emph{features}, so our trained network can automatically be reused by any segmentation model as long as it generates SAM-style object masks. These discrepancies make our method more flexible and feasible in real applications without needing labels.
\section{Method}

\begin{figure*}[tb]
  \centering
  \begin{subfigure}{0.3\linewidth}
    \includegraphics[width=\linewidth]{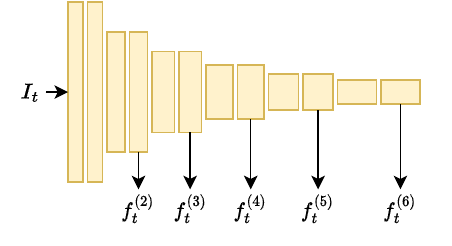}
    \caption{Encoder for each image $I_t$ ($t\in\{1, 2\}$)}
    \label{fig:enc}
  \end{subfigure}
  \hfill
  \begin{subfigure}{0.69\linewidth}
    \includegraphics[width=\linewidth]{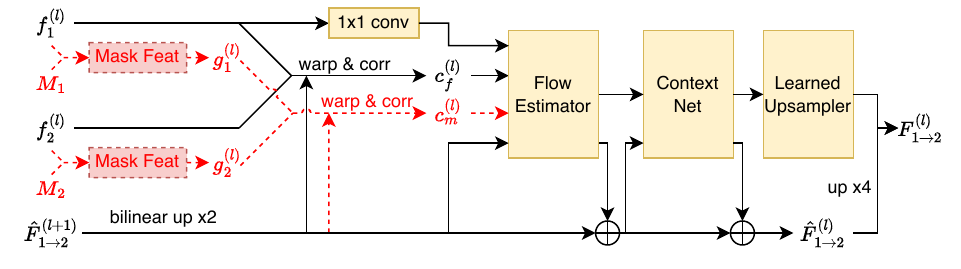}
    \caption{One iteration of the iterative decoder at the $l$-th level ($2\leq l\leq 6$).}
    \label{fig:dec}
  \end{subfigure}
\caption{Our network structure. The red part highlights our mask feature adaptation (``+mf''), which is only applied in our second setting where the SAM masks, $M_1$ and $M_2$, are used as additional inputs to the network. See more detailed network structures in Appendix A.1.}
\label{fig:network}
\end{figure*}

\begin{figure}
    \centering
    \includegraphics[width=\linewidth]{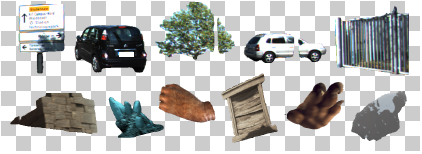}
    \caption{Examples of object crops selected from KITTI~\cite{kitti15} and Sintel~\cite{sintel}  using SAM~\cite{sam} for semantic augmentation (\cref{subsec:aug})}
    \label{fig:key_obj_demo}
\end{figure}

\subsection{Problem formulation} \label{subsec:prob_form}

\paragraph{Unsupervised optical flow}
Given two consecutive RGB frames $I_1, I_2\in\R^{H\times W\times 3}$, unsupervised optical flow estimation aims at estimating the dense optical flow field $F_{1\to 2}\in\R^{H\times W\times 2}$ without using ground-truth labels.

\paragraph{SAM mask detection}
For each input frame  $I_t$ ($t\in\{1, 2\}$), we can compute its SAM~\cite{sam} masks $M_t=\{0, 1\}^{n_t\times H\times W}$, which is composed of the binary masks of the $n_t$ objects found in $I_t$. The masks are generated by prompting SAM using a grid of around 1k points along with post-processing such as NMS~\cite{neubeck2006efficient} to drop redundant masks.
% \xiang{Explanation: ways to prompt the SAM mask, and why we choose this one}

To utilize SAM masks in optical flow training, one key question is how to effectively process, represent, and aggregate these masks in the network. Different from semantic or instance segmentation, SAM masks do not identify the semantic classes of each object, so one-hot representations are not applicable. Also, the number of detected masks $n_t$ may vary from sample to sample. Moreover, the relationship between these masks and all pixels is not strictly one-to-one. Instead, a pixel can belong to multiple different masks, for instance, in the case of embedded objects like cars and wheels. Conversely, some pixels may not be assigned to any mask at all. These technicalities have posted great challenges, and we will show how we tackle these issues in later chapters.

\paragraph{Two problem settings}
For the sake of practicality, we consider two settings. In the first setting, we do \emph{not} use SAM masks as additional inputs to the network. Instead, we only apply SAM during training to generate better loss signals, so SAM is not needed at inference time once training is completed. In addition, for the second setting, we also input SAM masks to the network to generate object-level mask features. This setting should yield better performances, albeit with the trade-off of extra overhead during the generation of SAM masks at inference time.

In this paper, we mainly propose three adaptations to utilize SAM information in the flow network, namely the semantic augmentation (\cref{subsec:aug}), the homography smoothness loss (\cref{subsec:hg}), and the mask feature module (\cref{subsec:mf}). These three adaptions can be plugged in independently. The former two only need SAM during training, so they can be applied to both problem settings. The last one involves adding SAM inputs and new learnable weights to the network, so we only apply it in the second setting.

\subsection{Network structure and loss}

We first build our baseline network from ARFlow~\cite{liu2020learning}, with some simple adaptations suggested by SemARFlow~\cite{yuan2023semarflow} such as adding the learned upsampler network. Our network structure is shown in \cref{fig:network}.

\paragraph{Encoder} We use a simple fully convolutional encoder (\cref{fig:enc}) to extract a feature pyramid $\{f_t^{(2)}, f_t^{(3)}, \cdots, f_t^{(6)}\}$ for each input image $I_t$ ($t\in\{1, 2\}$), where the $l$-th level feature $f_t^{(l)}$ has resolution $(H/2^{l}, W/2^l)$.

\paragraph{Decoder} We adopt the iterative decoder used in previous work~\cite{liu2020learning, yuan2023semarflow} as our decoder. The decoder starts from a coarse level zero estimate $\hat F_{1\to2}^{(7)} = 0$ and iteratively refines the estimate to finer levels. \cref{fig:dec} illustrates one iteration that refines from estimate $\hat F_{1\to2}^{(l+1)}$ to the finer $\hat F_{1\to2}^{(l)}$, which has resolution $(H/2^{l}, W/2^l)$. A learned upsampler network (similar to the one in RAFT~\cite{teed2020raft}) is applied to upsample $\hat F_{1\to2}^{(2)}$ by 4 times to generate our final flow estimate $F_{1\to2}=F_{1\to2}^{(2)}$ on the original resolution $(H, W)$.

In \cref{fig:dec}, we also highlight in red the optional mask feature module (to be discussed in \cref{subsec:mf}), which requires the SAM masks $M_1, M_2$ as additional inputs to the decoder and is thus only included in our second problem setting mentioned in \cref{subsec:prob_form}. See more details in Appendix A.2.

\paragraph{Loss} We adopt the same photometric loss $\ell_{\text{ph}}$ as in ARFlow~\cite{liu2020learning}, which is a linear combination of three distance measures ($\text{L}_1$, SSIM~\cite{wang2004image}, and Census loss~\cite{meister2018unflow}) between input frames and the frames warped by $F_{1\to2}$ and $F_{2\to1}$. Occluded regions estimated by bidirectional consistency check~\cite{meister2018unflow} are disregarded when computing $\ell_{\text{ph}}$.

In addition, we also combine a semantic augmentation loss $\ell_{\text{aug}}$ (\cref{subsec:aug}) and a homography smoothness loss $\ell_{\text{hg}}$ (\cref{subsec:hg}), so our final loss is 
\begin{equation} \label{eq:loss}
    \ell = \ell_{\text{ph}} + w_{\text{aug}}\ell_{\text{aug}} + w_{\text{hg}}\ell_{\text{hg}},
\end{equation}
where $w_{\text{aug}}=w_{\text{hg}}=0.1$ are the balancing weights.

\begin{figure*}
    \centering
    \includegraphics[width=\linewidth]{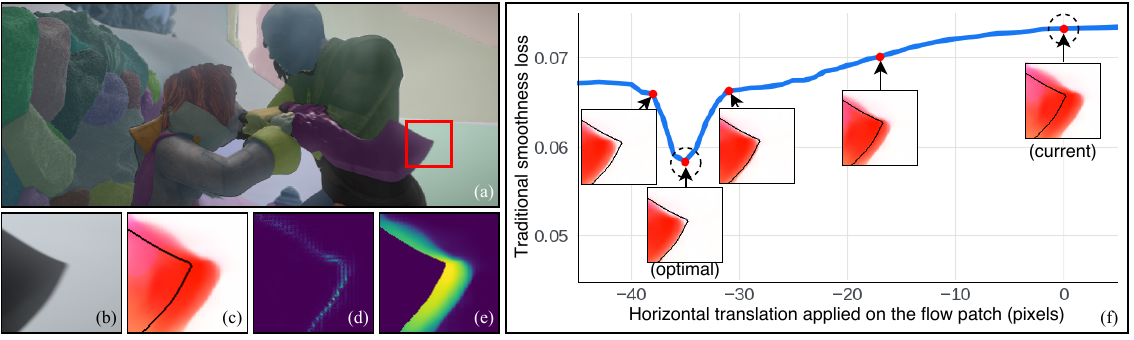}
    \caption{An example of why traditional boundary-aware smoothness loss works poorly. Sample from Sintel~\cite{sintel} final (ambush\_5, frame \#11). (a) Original image superimposed with SAM full segmentation; (b) Image patch; (c) Optical flow estimate from our baseline model superimposed with the SAM boundary (black); (d) Gradients of the traditional boundary-aware smoothness loss; (e) Gradients of our proposed homography smoothness loss; (f) Illustration of the poor landscape of traditional smoothness loss. Note that for both gradients in (d)(e), we use loss definitions based on $\text{L}_2$ norm for better visualizations. See \cref{subsec:hg} and Appendix A.6 for explanations.}
    \label{fig:boundary_smooth_issues}
\end{figure*}

\subsection{Semantic augmentation as self-supervision} \label{subsec:aug}

Inspired by SemARFlow~\cite{yuan2023semarflow}, we adopt a similar semantic augmentation process to improve our network by self-supervision during training. However, we extract semantics from SAM~\cite{sam} instead of semantic segmentation~\cite{zhu2019improving,chen2018encoder}.

\paragraph{Overview} 
After estimating the flow $F_{1\to2}$ for inputs $I_1$, $I_2$, we manually apply some transformations $\mathcal{T}_1$, $\mathcal{T}_2$ to $I_1$, $I_2$, respectively, to obtain the augmented $\tilde{I}_1$, $\tilde{I}_2$.  The new flow after transformation $\tilde F_{1\to2}$ can also be generated at the same time since all transformation parameters are known. We then run another forward pass of the network to infer flow for the augmented inputs $\tilde{I}_1$, $\tilde{I}_2$, which is then self-supervised by $\tilde F_{1\to2}$ using $\text{L}_1$ loss (\ie, the $\ell_\text{aug}$ in \cref{eq:loss}).

The transformations $\mathcal{T}_1$, $\mathcal{T}_2$ mentioned above not only include appearance transformations (on brightness, contrast, random noise, \etc), 2D affine transformations (translation, rotation, scaling), and occlusion augmentation (cropping), as proposed by ARFlow~\cite{liu2020learning}, but also contain a special semantic augmentation that involves input semantics, as proposed by SemARFlow~\cite{yuan2023semarflow}, which we discuss next.

\paragraph{Semantic augmentation}
During semantic augmentation, new objects are copied and pasted across samples. For example, we may crop out a car object from another random sample and paste it into the current sample used for training. An augmented simple motion is also applied to the cropped objects. This transformation utilizes the semantic knowledge and creates realistic samples with new occlusions.

In contrast to SemARFlow~\cite{yuan2023semarflow}, which picks object crops of specific classes such as cars and poles using semantic segmentation, our method utilizes SAM masks without class labels. Consequently, we select key objects among the SAM masks by finding those masks that overlap the most with other masks. This is based on the heuristic that key objects typically consist of multiple object parts that can also be detected by SAM. Some example key objects selected are shown in \cref{fig:key_obj_demo}. See more details in Appendix A.3.

\subsection{Homography smoothness loss} \label{subsec:hg}

Our second adaptation comes from the motivation that object segmentation can be used to formulate a more precise smoothness constraint to better regularize the optical flow field. 
We first analyze the issues of previous traditional smoothness losses and then show how we resolve those issues with the help of SAM~\cite{sam}.

\paragraph{Issues of previous smoothness losses} Most previous networks define their smoothness loss based on the second-order derivatives of the flow field~\cite{yu2016back}. Optical flow field $F_{1\to 2}$ is a two-dimensional function of point $\bm p=(x, y)$. Previous smoothness losses are typically in the form of 
\begin{equation} \label{eq:sm}
   \ell_{\text{s}} = \sum_{\bm p} \left(w_x(\bm p)\left\|\frac{\partial^2 F_{1\to2}(\bm p)}{\partial x^2}\right \| +w_y(\bm p)\left\|\frac{\partial^2 F_{1\to2} (\bm p)}{\partial y^2}\right \|\right),
\end{equation}
in which $w_x$, $w_y$ are the edge-aware weights to avoid penalties across object boundaries, where motion is not necessarily continuous. Such weights are usually derived from image edges, which often coincide with object boundaries~\cite{wang2018occlusion}. In our case, we can obtain more accurate boundaries from SAM masks. However, we find that these boundary-aware smoothness definitions work poorly.

One example is shown in \cref{fig:boundary_smooth_issues}. The patch (\cref{fig:boundary_smooth_issues}\hyperref[fig:boundary_smooth_issues]{b}) exhibits a rightward motion of the blade, occluding the nearby snow background, which barely moves. We show our baseline flow estimate, as well as the object boundary, in \cref{fig:boundary_smooth_issues}\hyperref[fig:boundary_smooth_issues]{c}. We can see that the estimated flow is not consistent with the object boundary due to occlusion (part of the snow regarded as moving together with the blade). In this case, the smoothness loss mostly comes from around the flow boundary, and so does its gradient (\cref{fig:boundary_smooth_issues}\hyperref[fig:boundary_smooth_issues]{d}). This gradient signal is very weak since the boundary only takes up a very smaller region, so its update is confined within only the small local neighborhood around the flow boundary.

Furthermore, we show that the landscape of the broadly used boundary-aware smoothness loss is problematic. We examine the smoothness loss of the patch while gradually translating the flow patch horizontally until it roughly fits the object boundary provided by SAM. The results are visualized in \cref{fig:boundary_smooth_issues}\hyperref[fig:boundary_smooth_issues]{f}. We can see that the optimal solution indeed finds the flow that is most consistent with the object boundary since we do not penalize across object boundaries. However, such solution lies in a very steep local minimum of the loss, while in contrast, the landscape around our current estimate is rather flat, meaning that any local change around the current estimate makes little difference to the loss. This vividly explains why traditional boundary-aware smoothness losses are very hard to optimize in training.

\paragraph{Regional smoothness based on homography}

Traditional boundary-aware smoothness (\cref{eq:sm}) works poorly since its definition and gradient are too local. To resolve this issue, our idea is to define smoothness based on object \emph{regions} instead of object \emph{boundaries}.

% \LL{
% Do we calculate the homograph between the initial flow estimation and SAM mask, and then warp the flow to match mask?Maybe we can be more clear in the description?}

Specifically, the inaccurate flow values in \cref{fig:boundary_smooth_issues}\hyperref[fig:boundary_smooth_issues]{c} can be understood as outliers in the same object region (snow). Thus, parametric models, such as homography, can be used to fix these outliers. For each object region of interest (found though occlusion estimation~\cite{meister2018unflow,yuan2022optical}), we first estimate its homography with RANSAC~\cite{fischler1981random} using the reliable correspondences provided by the current flow estimate. We define criteria to reject the estimated homography with low RANSAC inlier rate (see details in Appendix A.4). A refined flow can then be generated for that object using homography. We compute the $\text{L}_1$ distance between our current estimate and the refined flow as our homography smoothness loss $\ell_{\text{hg}}$ in \cref{eq:loss}. Our homography smoothness loss results in non-local gradients (\cref{fig:boundary_smooth_issues}\hyperref[fig:boundary_smooth_issues]{e}), which strongly enforces smoothness by regions.

One alternative, though, is to directly use the refined flow as our output, so the homography works through post-processing instead of loss signals~\cite{sevilla2016optical}. Nevertheless, we still prefer defining losses because in that case, homography and SAM are only needed during training. Empirically, we do not see big differences between their performances.

\subsection{Mask feature and correlation} \label{subsec:mf}
% We now aggregate masks and features in the network.

\paragraph{Full segmentation representation of SAM masks} To better use masks in the network, we need to transform SAM masks to a full segmentation representation, where every pixel is assigned to exactly one mask.

Specifically, we first sort all current object masks by the size of their area. For pixels that belong to multiple masks, we assign it to the one that has the smallest area. For pixels that do not belong to any mask, we create a new ``background'' object mask to cover all these pixels.

\begin{figure}[tb]
    \centering
    \includegraphics[width=\linewidth]{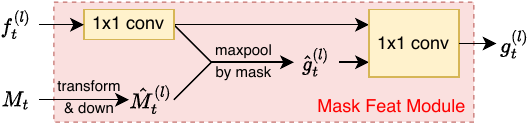}
    \caption{Our proposed mask feature module (\cref{subsec:mf})}
    \label{fig:mf}
\end{figure}

\paragraph{Mask feature module}
Our mask feature computations are highlighted in \cref{fig:dec}. Given the mask $M_t$ and image feature $f_t^{(l)}$, we use a mask feature module to generate $g_t^{(l)}$. Similar to the warping and correlation of image features $f_t^{(l)}$ in the original ARFlow~\cite{liu2020learning}, we also warp mask features $g_t^{(l)}$ and compute their correlations, which are then concatenated into the input of flow estimator network in \cref{fig:dec}.

The detailed structure of our mask feature module is depicted in \cref{fig:mf}. We first transform the SAM masks $M_t$ to full segmentation representation and downsize it to the $l$-level resolution. We then compute a new feature from $f_t^{(l)}$ and apply max-pooling by segmentation. Specifically, for each object segmentation, we apply max-pooling among all features of that object and copy the pooled feature to all those pixels, yielding a pooled feature map $\hat g_t^{(l)}$. Finally, we concatenate the feature before max-pooling with $\hat g_t^{(l)}$, and extract our mask feature $g_t^{(l)}$. This module aggregates features on the higher object level and can, thus, compensate the original pixel-level image features. See more details in Appendix A.5.

\section{Experiments}

\begin{table*}[tb]
\centering
\begin{tabular}{cl|cc|cccccc|c}
\thickhline
\multicolumn{2}{c|}{\multirow{3}{*}{Method}}   &\multicolumn{2}{c|}{Train}              & \multicolumn{6}{c|}{Test}            & \multicolumn{1}{c}{\multirow{3}{*}{Param.}}                           \\ \cline{3-10}
\multicolumn{2}{c|}{}                        & \multicolumn{1}{c|}{2012} & \multicolumn{1}{c|}{2015} & \multicolumn{2}{c|}{2012}         & \multicolumn{4}{c|}{2015}     &    \\
\multicolumn{2}{c|}{}                      & \multicolumn{1}{c|}{EPE}  & EPE             & \underline{Fl-noc} & \multicolumn{1}{c|}{EPE} & \underline{Fl-all} & Fl-noc & Fl-bg & Fl-fg & \\ \hline
\multicolumn{1}{c|}{\multirow{5}{*}{\rotatebox[origin=c]{90}{Supervised}}}     & PWC-Net+ ~\cite{sun2019models}         & \multicolumn{1}{c|}{-}     & (1.50)  &  3.36    & \multicolumn{1}{c|}{1.4}    & 7.72  &  4.91    &   7.69   & 7.88    & 8.8M \\
\multicolumn{1}{c|}{}                         & IRR-PWC~\cite{hur2019iterative}           & \multicolumn{1}{c|}{-}    & (1.63)     & 3.21 & \multicolumn{1}{c|}{1.6} & 7.65  & 4.86 & 7.68 & 7.52 & 6.4M \\
\multicolumn{1}{c|}{}                           & RAFT~\cite{teed2020raft}              & \multicolumn{1}{c|}{-}    & (0.63)     & -    & \multicolumn{1}{c|}{-}   & 5.10  & 3.07 & 4.74 & 6.87 & 5.3M \\
\multicolumn{1}{c|}{}                           & FlowFormer~\cite{huang2022flowformer}    & \multicolumn{1}{c|}{-}    & (0.53) & -    & \multicolumn{1}{c|}{-}   & 4.68  & \textbf{2.69} & \textbf{4.37} & \textbf{6.18} & 18.2M \\ 
\multicolumn{1}{c|}{}                           & SAMFlow~\cite{zhou2023samflow}$^{*\dagger}$        & \multicolumn{1}{c|}{-}    & -     & -    & \multicolumn{1}{c|}{-}   & \textbf{4.49} & - & - & - &  -\\ \hline
\multicolumn{1}{c|}{\multirow{11}{*}{\rotatebox[origin=c]{90}{Unsupervised}}} & UnFlow-CSS~\cite{meister2018unflow}        & \multicolumn{1}{c|}{3.29} & 8.10                     & -   & \multicolumn{1}{c|}{-} & 23.27  & -   & -  & -  & 116.6M\\  % 116.58M
% \multicolumn{1}{c|}{}                          & DSTFlow~\cite{ren2017unsupervised}            & \multicolumn{1}{c|}{10.43} & 16.79                &  -  & \multicolumn{1}{c|}{12.40} & 39.00  &  -  & - & - \\
\multicolumn{1}{c|}{}                          & DDFlow~\cite{liu2019ddflow}      & \multicolumn{1}{c|}{2.35} & 5.72                     & 4.57   & \multicolumn{1}{c|}{3.0} & 14.29  & 9.55   & 13.08	 & 20.40  & 4.3M \\  % 4.27M
\multicolumn{1}{c|}{}                          & SelFlow~\cite{liu2019selflow}          & \multicolumn{1}{c|}{1.69} & 4.84                     & 4.31   & \multicolumn{1}{c|}{2.2} & 14.19  & 9.65   & 12.68 & 21.74 & 4.8M \\  % 4.79M
\multicolumn{1}{c|}{}                          & SimFlow~\cite{im2020unsupervised}           & \multicolumn{1}{c|}{-} & 5.19                     & -   & \multicolumn{1}{c|}{-} & 13.38  & 8.21   & 12.60 & 17.27 & - \\
\multicolumn{1}{c|}{}                           & ARFlow ~\cite{liu2020learning}          & \multicolumn{1}{c|}{1.44} & 2.85       & 5.02    & \multicolumn{1}{c|}{1.8} & 11.80 & 8.91    & 10.30    & 19.32   & 2.2M \\  % 2.24M
\multicolumn{1}{c|}{}                           & UFlow ~\cite{jonschkowski2020matters}            & \multicolumn{1}{c|}{1.68} & (2.71)       & 4.26 & \multicolumn{1}{c|}{1.9} & 11.13 & 8.41 & 9.78 & 17.87 & - \\
\multicolumn{1}{c|}{}                           & UPFlow ~\cite{luo2021upflow}          & \multicolumn{1}{c|}{{1.27}} & 2.45      & -    & \multicolumn{1}{c|}{{\textbf{1.4}}} & 9.38  & -    & -    & -   & 3.5M \\ \cline{2-11}   % 3.49M
\multicolumn{1}{c|}{}                           & Ours (baseline)  & \multicolumn{1}{c|}{ 1.32} &   2.44                   &  4.05   & \multicolumn{1}{c|}{1.6 } & 9.60  & 6.77   & 8.74 & \textbf{13.89} & 2.5M \\  % baseline/20230622_110520_kitti_baseline_f450860686  2.513M
\multicolumn{1}{c|}{}                         & Ours (+aug)$^{*}$ & \multicolumn{1}{c|}{1.33} & 2.26                    & 4.15   & \multicolumn{1}{c|}{1.6} & 9.05 & 6.46   & 7.96  & 14.55 & 2.5M \\  % aug/20230731_161260_kitti_aug01
\multicolumn{1}{c|}{}                         & Ours (+aug +hg)$^{*}$ & \multicolumn{1}{c|}{1.27} & 2.11                    & 3.89   & \multicolumn{1}{c|}{1.5} & 8.18 & 6.04   & 6.67  & 15.72 & 2.5M \\  % hg/20230731_165060_kitti_aug01_hg01
\multicolumn{1}{c|}{}                         & Ours (+aug +hg +mf)$^{*\dagger}$   & \multicolumn{1}{c|}{\textbf{1.26}} & \textbf{2.01}                   & \textbf{3.79}   & \multicolumn{1}{c|}{\textbf{1.4}} & \textbf{7.83} &  \textbf{5.67}  & \textbf{6.40}  & 14.98 & 2.6M \\   % mask_corr/20230807_121160_kitti_concat_aug_hg  2.625M
\thickhline
\end{tabular}
\caption{KITTI benchmark errors (EPE/px and Fl/\%). Metrics evaluated at ``all'' (all pixels), ``noc'' (non-occlusions), ``bg'' (background), and ``fg'' (foreground). ``+aug'': semantic augmentation module; ``+hg'': homography smoothness loss; ``+mf'': mask feature module. ``*'': SAM used in training; ``$\dagger$'': SAM used in inference. Numbers with parentheses indicate that the same evaluation data were used in training.}
%Our `baseline' is an adapted ARFlow with added learned upsampler and no smoothness loss. `+enc` means adding semantic encoder; `+aug' means adding semantic augmentation.  `-' means unavailable. $^\dagger$ denotes models with semantic inputs. 
\label{tab:unsup_test_kitti}
\end{table*}

\begin{table*}[tb]
\centering
\begin{tabular}{cl|cc|cccccc|c}
\thickhline
\multicolumn{2}{c|}{\multirow{3}{*}{Method}}  & \multicolumn{2}{c|}{Train}              & \multicolumn{6}{c|}{Test}             & \multicolumn{1}{c}{\multirow{3}{*}{Param.}}                            \\ \cline{3-10}
\multicolumn{2}{c|}{}                       & \multicolumn{1}{c|}{Clean} & \multicolumn{1}{c|}{Final} & \multicolumn{3}{c|}{Clean}         & \multicolumn{3}{c|}{Final}   &     \\
\multicolumn{2}{c|}{}                       & \multicolumn{1}{c|}{all}  & all             & \underline{all} & noc & \multicolumn{1}{c|}{occ} & \underline{all} & noc & occ  & \\ \hline
\multicolumn{1}{c|}{\multirow{5}{*}{\rotatebox[origin=c]{90}{Supervised}}}     & PWC-Net+ ~\cite{sun2019models}       & \multicolumn{1}{c|}{(1.71)}    & (2.34) & 3.45   & 1.41 & \multicolumn{1}{c|}{20.12}   & 4.60 & 2.25 & 23.70 & 8.8M \\ 
\multicolumn{1}{c|}{}                         & IRR-PWC~\cite{hur2019iterative}      & \multicolumn{1}{c|}{(1.92)}    & (2.51) & 3.84  & 1.47  & \multicolumn{1}{c|}{23.22}   & 4.58  & 2.15  & 24.36 & 6.4M \\ 
\multicolumn{1}{c|}{}                           & RAFT~\cite{teed2020raft}      & \multicolumn{1}{c|}{(0.77)}      & (1.27)        & {1.61}  & {0.62}  & \multicolumn{1}{c|}{9.65}   & {2.86}  & {1.41}  & {14.68} & 5.3M \\
\multicolumn{1}{c|}{}                           & FlowFormer~\cite{huang2022flowformer}    & \multicolumn{1}{c|}{(0.48)}    & (0.74) & 1.16  & 0.42   & \multicolumn{1}{c|}{7.16}   & 2.09 & \textbf{0.96} & 11.30 & 18.2M \\ 
\multicolumn{1}{c|}{}                           & SAMFlow~\cite{zhou2023samflow}$^{*\dagger}$        & \multicolumn{1}{c|}{-}    & - &  \textbf{1.00}  & \textbf{0.38} & \multicolumn{1}{c|}{\textbf{5.97}}   & \textbf{2.08} & 1.04 & \textbf{10.60} & - \\ \hline
\multicolumn{1}{c|}{\multirow{11}{*}{\rotatebox[origin=c]{90}{Unsupervised}}} &  UnFlow-CSS~\cite{meister2018unflow}              & \multicolumn{1}{c|}{-}  & 7.91  & 9.38 & 5.37 & \multicolumn{1}{c|}{42.11} & 10.22 & 6.06 & 44.11 & 116.6M \\ 
 % &  DSTFlow~\cite{ren2017unsupervised}                  & \multicolumn{1}{c|}{(6.16)}  & (7.38)  & 10.41 & 5.30 & \multicolumn{1}{c|}{-} & 11.28 & 6.16 & -  \\ 
\multicolumn{1}{c|}{}                          &  DDFlow~\cite{liu2019ddflow}             & \multicolumn{1}{c|}{(2.92)}  & (3.98)  & 6.18 & 2.27 & \multicolumn{1}{c|}{38.05} & 7.40 & 3.41 & 39.94 & 4.3M \\  
\multicolumn{1}{c|}{}                          & SelFlow~\cite{liu2019selflow}                  & \multicolumn{1}{c|}{(2.88)}  & (3.87)  & 6.56 & 2.67 & \multicolumn{1}{c|}{38.30} & 6.57 & 3.12 & 34.72 & 4.8M \\ 
\multicolumn{1}{c|}{}                          & SimFlow~\cite{im2020unsupervised}      & \multicolumn{1}{c|}{(2.86)}    & (3.57) & 5.93   & 2.16 & \multicolumn{1}{c|}{36.66}   & 6.92 & 3.02 & 38.70 & - \\ 
\multicolumn{1}{c|}{}                           & ARFlow ~\cite{liu2020learning}         & \multicolumn{1}{c|}{(2.79)}         & (3.73)        & {4.78}  & {1.91}  & \multicolumn{1}{c|}{28.26}  & {5.89}  & {2.73}  & {31.60} & 2.2M \\ 
\multicolumn{1}{c|}{}                           & UFlow ~\cite{jonschkowski2020matters}                 & \multicolumn{1}{c|}{(2.50)}          & (3.39)        & 5.21  & 2.04  & \multicolumn{1}{c|}{31.06}  & 6.50  & 3.08  & 34.40 & - \\ 
\multicolumn{1}{c|}{}                           & UPFlow ~\cite{luo2021upflow}            & \multicolumn{1}{c|}{(2.33)}    & (2.67) &  4.68  & 1.71 & \multicolumn{1}{c|}{28.95}   & 5.32 & \textbf{2.42} & 28.93  & 3.5M \\ \cline{2-11} 
\multicolumn{1}{c|}{}                           & Ours (baseline)       & \multicolumn{1}{c|}{(2.67)}    & (3.63) & 4.29   & {1.64} & \multicolumn{1}{c|}{25.96}   & 5.81 & 2.76 & 30.60 & 2.5M \\  % baseline/20230628_140130_sintel_baseline_f453056634
\multicolumn{1}{c|}{}                         & Ours (+aug)$^{*}$       & \multicolumn{1}{c|}{(2.35)}    & (3.33) & 4.00   & \textbf{1.58} & \multicolumn{1}{c|}{23.76}   & 5.33 & 2.53 & 28.17 & 2.5M \\ % aug/20231019_110816_sintel_aug_f492918489
\multicolumn{1}{c|}{}                         & Ours (+aug +hg)$^{*}$       & \multicolumn{1}{c|}{(2.25)}    & (3.10) & 4.00   & 1.76 & \multicolumn{1}{c|}{22.36}   & 5.22 & 2.62 & \textbf{26.40} & 2.5M \\   % hg/20230802_095260_sintel_all_aug01_hg01
\multicolumn{1}{c|}{}                         & Ours (+aug +hg +mf)$^{*\dagger}$          & \multicolumn{1}{c|}{(2.21)}    & (3.07) &  \textbf{3.93}  & 1.67 & \multicolumn{1}{c|}{\textbf{22.34}}   & \textbf{5.20} & 2.56 & 26.75 & 2.6M \\  % mask_corr/20231016_135412_sintel_aug+hg+mf_f491850254
\thickhline
\end{tabular}
\caption{Sintel benchmark errors (EPE/px). Metrics evaluated at ``all'' (all pixels), ``noc'' (non-occlusions), and ``occ'' (occlusions). ``+aug'': semantic augmentation module; ``+hg'': homography smoothness loss; ``+mf'': mask feature module. ``*'': SAM used in training; ``$\dagger$'': SAM used in inference.  Numbers with parentheses indicate that the same evaluation data were used in training.}
\label{tab:unsup_test_sintel}
\end{table*}

% Our final test results vs. other semantic flow results
\begin{table}[tb]
\centering
\begin{tabular}{lc|cc}
\thickhline
\multicolumn{1}{c}{\multirow{2}{*}{Method}} & \multirow{2}{*}{Semantics} &  \multicolumn{2}{c}{KITTI}  \\ 
\multicolumn{1}{c}{}  & \multicolumn{1}{c|}{}   & 2015 & 2012 \\ \hline
JFS~\cite{hur2016joint}       & Sem. Seg.                   & 16.47 & - \\
SOF~\cite{sevilla2016optical}    & Sem. Seg.               & 15.99 & -  \\
MRFlow~\cite{wulff2017optical}     & Sem. Seg.              & 12.19 &  -  \\  
SDF\cite{bai2016exploiting}        & Ins. Seg.               & 11.01  & 7.69  \\
SemARFlow~\cite{yuan2023semarflow}     & Sem. Seg.                  & 8.38   & 7.35 \\ \hline
Ours (+aug +hg +mf)  & SAM    & \textbf{7.83} & \textbf{7.05} \\ 
\thickhline
\end{tabular}
\caption{KITTI test errors (Fl-all/\%) compared with other unsupervised semantic optical flow methods. ``-'': data not available.}
\label{tab:sem_flow_test}
\end{table}

\subsection{Datasets}

We conduct experiments on KITTI~\cite{kitti12,kitti15} and Sintel~\cite{sintel} datasets and follow the same training data schedules from previous methods~\cite{liu2019selflow,liu2020learning,luo2022learning}. For KITTI~\cite{kitti12,kitti15}, we first train on raw sequences (55.7k samples) and then fine-tune on the multi-view extension subset (5.9k samples). For Sintel~\cite{sintel}, we first train on raw frames (12.5k samples) provided by ARFlow~\cite{liu2020learning} and then fine-tune on clean and final passes together (2.1k samples). We also adopt the Sintel sub-splits used in ARFlow~\cite{liu2020learning}, which divide the original dataset by scenes into two subsets of 1k samples, for two-fold cross validation. Images from the test scenes have been excluded from the raw sequences for both datasets.

\subsection{Implementation details}

The model is implemented in PyTorch~\cite{NEURIPS2019_9015}. We train the network using the Adam optimizer~\cite{kingma2014adam} ($\beta_1=0.9, \beta_2=0.999$) with batch size 8. For both datasets, we first train on raw data using a constant learning rate 2e-4 for 100k iterations and then fine-tune on the original dataset using the OneCycleLR scheduler~\cite{smith2019super} with maximum learning rate 4e-4 for another 100k iterations. Similar to SemARFlow~\cite{yuan2023semarflow}, we only turn on the semantic augmentation and homography smoothness modules after 150k iterations.

In terms of Segment Anything Model~\cite{sam}, we use the off-the-shelf default ViT-H pretrained model, which generates an average of 63.7 object masks for each KITTI sample~\cite{kitti15} and around 82.9 masks for each Sintel sample~\cite{sintel}.

For data augmentation, we follow ARFlow~\cite{liu2020learning} and include appearance transformations (brightness, contrast, saturation, hue, gaussian blur, \etc), random horizontal flipping, and random swapping of input images. We resize the inputs to dimension $256\times832$ for KITTI and $448\times1024$ for Sintel before feeding into the network.

% For data augmentation, we include random horizontal flipping and swapping of the input frames. We resize the inputs to $256\times 832$ before feeding into the network. The photometric loss weight for each scale $a_l$ $(2\leq l\leq 6)$ in \cref{eq:ph_loss_one_scale} are set as 1, 1, 1, 1, 0. The weights for three photometric distance measures $\rho_k$ $(1\leq k\leq 3)$ in \cref{eq:ph_loss_one_scale} are set as 0.15, 0.85, 0 for the first 50k iterations and 0, 0, 1 afterwards. We start the appearance and spatial augmentation (second pass as in ARFlow~\cite{liu2020learning}) at 50k iterations, and semantic augmentation (third pass) after 150k iterations. 

\subsection{Benchmark testing} \label{subsec:benchmark}

Our benchmark testing results are shown in \cref{tab:unsup_test_kitti,tab:unsup_test_sintel}. Our final models with all three adaptations significantly outperform state-of-the-art unsupervised methods on both KITTI~\cite{kitti12,kitti15} and Sintel~\cite{sintel} datasets on almost all evaluation metrics. Our final model achieves 7.84\% error rate on KITTI-2015 test set, which is much better than UPFlow~\cite{luo2022learning} (9.38\%) and ARFlow~\cite{liu2020learning} (11.80\%, the backbone network that we adapt from). All these results show the benefits of utilizing SAM models in unsupervised optical flow training.

In \cref{tab:unsup_test_kitti,tab:unsup_test_sintel}, we can also see that our errors decline progressively as we incrementally add each proposed module to the network. This justifies the effectiveness of all our proposed adaptations. Also, for the setting that we do not use SAM masks as network inputs (``+aug +hg''), our model also outperforms state-of-the-art methods on both datasets. This implies that our approach has the potential to enhance unsupervised flow networks solely by optimizing training, guided by SAM, without introducing any additional computational overhead during inference.

Notably, our networks exhibit substantial improvements from SAM particularly on real datasets, such as KITTI~\cite{kitti12,kitti15}, compared with animation images in Sintel~\cite{sintel}. This is because SAM is mostly trained on real-life images, so it produces masks of higher quality for KITTI than for Sintel.

\cref{tab:sem_flow_test} shows the comparison among current semantics-guided optical flow methods. Our model guided by SAM outshines all previous methods guided by semantic or instance segmentation, even though SAM is not trained on KITTI~\cite{kitti15}. This indicates the great potential of SAM as a zero-shot general-purpose semantic model that could be used directly in other tasks such as optical flow estimation.

\subsection{Qualitative results} \label{subsec:qual}

\begin{figure*}
    \centering
    \includegraphics[width=\linewidth]{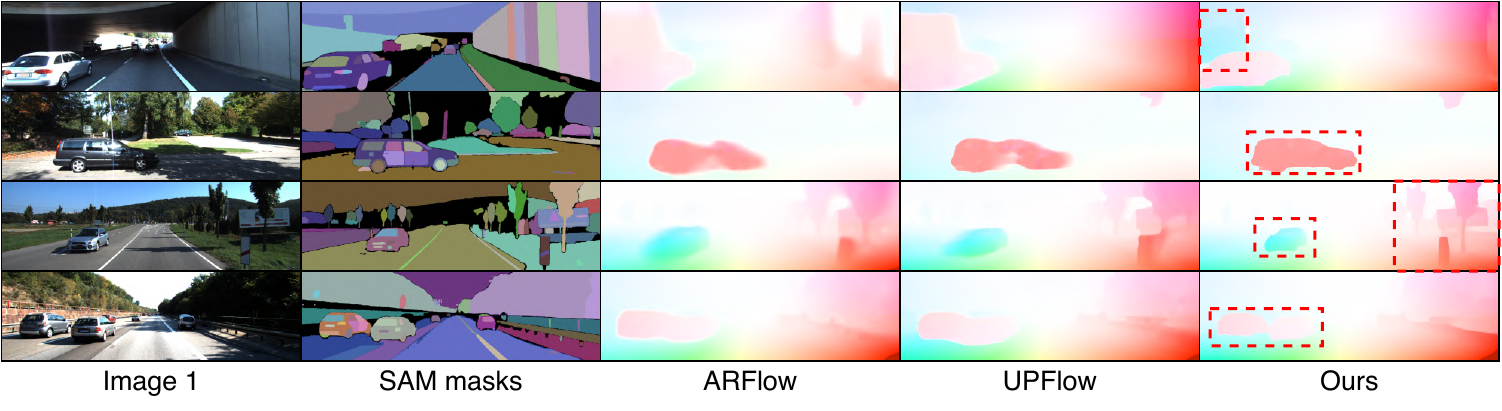}
    \caption{KITTI-2015 test qualitative examples (sample frame \#48, 72, 190, 196). See more examples in Appendix B.2}
    \label{fig:kitti_examples}
\end{figure*}

\begin{figure*}
    \centering
    \includegraphics[width=\linewidth]{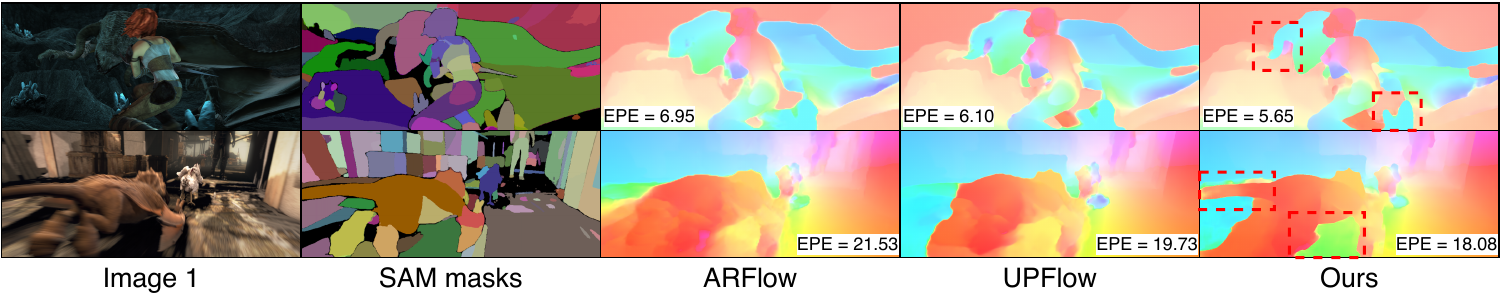}
    \caption{Sintel (final pass) test qualitative examples (sample: cave\_3 frame 16; market\_4 frame 47). See more examples in Appendix B.2}
    \label{fig:sintel_examples}
\end{figure*}

\cref{fig:kitti_examples,fig:sintel_examples} show some qualitative examples of our final model, compared with previous state-of-the-art methods. We can see that our network outputs better flow around objects with much sharper boundaries, which are consistent with the SAM mask inputs. Our method can also handle different lighting conditions (dark shadows, bright reflections) better thanks to the robust masks provided by SAM.

%% kitti examples
% 9: many cars
% 16: car and poles
% 20: poles occlusion
% 44: many cars
% 48*: black confusing part
% 63: good car and tree
% 72*: lighting good car
% 76, 157: clear car
% 78, 183: clear sky
% 134, 194: good car
% 165: clear poles
% 190*: good objects
% 196*: separate car

\subsection{Ablation studies} \label{subsec:ablate}

We do extensive ablation studies to analyze the effectiveness of our proposed adaptations and their detailed settings. For KITTI~\cite{kitti12,kitti15}, we compute validation errors on the original train set images. For Sintel~\cite{sintel}, we apply two-fold cross validation and report the average validation errors.

\begin{table}[tb]
\centering
\begin{tabular}{cc|cc|cc}
\thickhline
\multirow{2}{*}{$w_{\text{aug}}$} & \multirow{2}{*}{$w_\text{hg}$} & \multicolumn{2}{c|}{Sintel} & \multicolumn{2}{c}{KITTI} \\
                        &                        & Final        & Clean        & 2015        & 2012        \\ \hline
0.1*                     & 0.1*                    & \textbf{3.35}         & \textbf{2.53}         & 2.11        & \textbf{1.27}        \\ \hline
0.1                     & 0                      &     3.53         &    2.54          & 2.26        & 1.33        \\
0.1                     & 0.2                    & 3.45         & 2.63         &  \textbf{2.09}           &   1.28          \\
0.2                     & 0.2                    & 3.45         & 2.60          & 2.15        & 1.29        \\
0.2                     & 0.1                    & 3.50          & {2.59}         & 2.13        & 1.29        \\
0                       & 0.1                    & 3.61         & 2.79         & 2.16        & 1.30         \\ 
\thickhline
\end{tabular}
\caption{Ablation Study on Ours (+aug +hg): Balancing weights between $w_\text{aug}$ and $w_\text{hg}$ in \cref{eq:loss}. * indicates our final setting.}
\label{tab:ablation_1}
\end{table}

\begin{table}[tb]
\centering
\begin{tabular}{c|cc|cc}
\thickhline
\multirow{2}{*}{Smoothness Loss}  & \multicolumn{2}{c|}{Sintel} & \multicolumn{2}{c}{KITTI} \\
                                               & Final        & Clean        & 2015        & 2012        \\ \hline
Homography*                                       & \textbf{3.35}         & \textbf{2.53}         & \textbf{2.11}        & \textbf{1.27}        \\ \hline
Image-edge-aware                                     &      3.63       &     {2.56}         & 2.35        & 1.35        \\
SAM-boundary-aware                                 &   3.65       &   2.62      &     2.33        &    1.36         \\
\thickhline
\end{tabular}
\caption{Ablation Study on Ours (+aug +hg): Different smoothness loss definition (\cref{subsec:hg}). * indicates our current setting.}
\label{tab:ablation_2}
\end{table}

\begin{table}[tb]
\centering
\begin{tabular}{c|cc|cc}
\thickhline
\multirow{2}{*}{Mask Feature Module}  & \multicolumn{2}{c|}{Sintel} & \multicolumn{2}{c}{KITTI} \\
                                               & Final        & Clean        & 2015        & 2012        \\ \hline
Concat*                                       & \textbf{3.29}         & \textbf{2.43}         & \textbf{2.01}        & \textbf{1.26}        \\ \hline
Residual                                    &   3.31          &  2.47            & 2.05        & 1.28        \\
Concat + Residual                                &   3.34       &   2.52      &     2.06        &    1.27        \\
No +mf  &    3.35     &    2.53     &     2.11        &    1.27         \\
\thickhline
\end{tabular}
\caption{Ablation Study on Ours (+aug +hg +mf): Different mask feature modules (\cref{subsec:mf}).  * indicates our final setting.}
\label{tab:ablation_3}
\end{table}

\paragraph{Loss weights} We tune the loss weights $w_{\text{aug}}$ and $w_\text{hg}$ in \cref{tab:ablation_1}. We also compare with the settings where either $w_{\text{aug}}$ or $w_\text{hg}$ equals zero, which means we turn off the semantic augmentation or homography smoothness module. The results show that our current setting works the best, and the ablation of each module results in loss of performance.

\paragraph{Smoothness definitions} In \cref{tab:ablation_2}, we can see that our regional smoothness loss based on homography works significantly better than the traditionally used edge-aware smoothness loss based on image edges or SAM boundaries. These results are consistent with our analysis in \cref{subsec:hg}.

\paragraph{Mask feature modules} We also experiment some other ways of aggregating mask features and image features in \cref{tab:ablation_3}. ``Residual'' refers to adding the processed mask features to image features as a residual connection. The results show that our current setting works slightly better than using residual connections.

\subsection{Generalization ability} \label{subsec:generalize}

We show that our flow network guided by SAM exhibits great generalization ability across dataset domains in \cref{tab:generalization}. Specifically, we train on one of the datasets (KITTI~\cite{kitti15} or Sintel~\cite{sintel}) and then test directly on the other without fine-tuning. Our final model guided by SAM obtains clearly better results than our baseline model without SAM.

\setlength{\abovecaptionskip}{5pt}
\setlength{\belowcaptionskip}{-5pt}
\begin{table}[tb]
\centering
\begin{tabular}{c|cc|cc}
\thickhline
\multirow{2}{*}{Method}  & \multicolumn{2}{c|}{KITTI$\to$Sintel} & \multicolumn{2}{c}{Sintel$\to$KITTI} \\
                                               & Final        & Clean        & 2015        & 2012        \\ \hline
Ours (w/ SAM)                                       & \textbf{5.75}         & \textbf{4.90}         &  \textbf{7.58}        &  2.99       \\ 
Ours (w/o SAM)                                     &  7.02           &  6.39            & 8.59        & \textbf{2.93}        \\
\thickhline
\end{tabular}
\caption{Generalization ability. Training on one dataset and testing directly on the other dataset. We show Sintel/KITTI train set EPEs.}
\label{tab:generalization}
\end{table}
\setlength{\abovecaptionskip}{10pt}
\setlength{\belowcaptionskip}{0pt}

\subsection{Time efficiency} \label{subsec:time}

Our network operates very efficiently in real time. For each RGB sample of dimension $376\times1242$, our network inference takes 0.0334$(\pm0.0038)$ second on one Tesla P100 GPU,
excluding the time for computing SAM masks.

\section{Conclusion}

We propose UnSAMFlow, an unsupervised optical flow network guided by object information from Segment Anything Model (SAM), with three novel adaptations, namely semantic augmentation, homography smoothness, and mask feature correlation. Our method achieves state-of-the-art results and exhibits visible improvements. 

\paragraph{Limitations} Our performance relies on the accuracy of SAM masks, which may be undermined for samples with serious lighting issues, artifacts, or motion blur. The lack of semantic classes in the SAM output also makes its object information incomplete, awaiting future improvements.

\newpage
{
    \small
    \bibliographystyle{format_cvpr2024/ieeenat_fullname}
    \bibliography{refs}

\begin{thebibliography}{75}
\providecommand{\natexlab}[1]{#1}
\providecommand{\url}[1]{\texttt{#1}}
\expandafter\ifx\csname urlstyle\endcsname\relax
  \providecommand{\doi}[1]{doi: #1}\else
  \providecommand{\doi}{doi: \begingroup \urlstyle{rm}\Url}\fi

\bibitem[Bai et~al.(2016)Bai, Luo, Kundu, and Urtasun]{bai2016exploiting}
Min Bai, Wenjie Luo, Kaustav Kundu, and Raquel Urtasun.
\newblock Exploiting semantic information and deep matching for optical flow.
\newblock In \emph{ECCV}, pages 154--170. Springer, 2016.

\bibitem[Butler et~al.(2012)Butler, Wulff, Stanley, and Black]{sintel}
Daniel~J. Butler, Jonas Wulff, Garrett~B. Stanley, and Michael~J. Black.
\newblock A naturalistic open source movie for optical flow evaluation.
\newblock In \emph{ECCV}, pages 611--625. Springer-Verlag, 2012.

\bibitem[Cen et~al.(2023)Cen, Zhou, Fang, Shen, Xie, Zhang, and Tian]{cen2023segment}
Jiazhong Cen, Zanwei Zhou, Jiemin Fang, Wei Shen, Lingxi Xie, Xiaopeng Zhang, and Qi Tian.
\newblock Segment anything in 3d with nerfs.
\newblock \emph{arXiv preprint arXiv:2304.12308}, 2023.

\bibitem[Chen et~al.(2018)Chen, Zhu, Papandreou, Schroff, and Adam]{chen2018encoder}
Liang-Chieh Chen, Yukun Zhu, George Papandreou, Florian Schroff, and Hartwig Adam.
\newblock Encoder-decoder with atrous separable convolution for semantic image segmentation.
\newblock In \emph{ECCV}, pages 801--818, 2018.

\bibitem[Chen et~al.(2023)Chen, Tang, Wan, Wang, and Zeng]{chen2023interactive}
Xiaokang Chen, Jiaxiang Tang, Diwen Wan, Jingbo Wang, and Gang Zeng.
\newblock Interactive segment anything nerf with feature imitation.
\newblock \emph{arXiv preprint arXiv:2305.16233}, 2023.

\bibitem[Cheng et~al.(2017)Cheng, Tsai, Wang, and Yang]{cheng2017segflow}
Jingchun Cheng, Yi-Hsuan Tsai, Shengjin Wang, and Ming-Hsuan Yang.
\newblock Segflow: Joint learning for video object segmentation and optical flow.
\newblock In \emph{ICCV}, pages 686--695, 2017.

\bibitem[Cheng et~al.(2023)Cheng, Li, Xu, Li, Yang, Wang, and Yang]{cheng2023segment}
Yangming Cheng, Liulei Li, Yuanyou Xu, Xiaodi Li, Zongxin Yang, Wenguan Wang, and Yi Yang.
\newblock Segment and track anything.
\newblock \emph{arXiv preprint arXiv:2305.06558}, 2023.

\bibitem[Ding et~al.(2020)Ding, Wang, Zhou, Shi, Lu, and Luo]{ding2020every}
Mingyu Ding, Zhe Wang, Bolei Zhou, Jianping Shi, Zhiwu Lu, and Ping Luo.
\newblock Every frame counts: Joint learning of video segmentation and optical flow.
\newblock In \emph{AAAI}, pages 10713--10720, 2020.

\bibitem[Dosovitskiy et~al.(2015)Dosovitskiy, Fischer, Ilg, Hausser, Hazirbas, Golkov, Van Der~Smagt, Cremers, and Brox]{dosovitskiy2015flownet}
Alexey Dosovitskiy, Philipp Fischer, Eddy Ilg, Philip Hausser, Caner Hazirbas, Vladimir Golkov, Patrick Van Der~Smagt, Daniel Cremers, and Thomas Brox.
\newblock Flownet: Learning optical flow with convolutional networks.
\newblock In \emph{ICCV}, pages 2758--2766, 2015.

\bibitem[Dosovitskiy et~al.(2021)Dosovitskiy, Beyer, Kolesnikov, Weissenborn, Zhai, Unterthiner, Dehghani, Minderer, Heigold, Gelly, Uszkoreit, and Houlsby]{dosovitskiy2020image}
Alexey Dosovitskiy, Lucas Beyer, Alexander Kolesnikov, Dirk Weissenborn, Xiaohua Zhai, Thomas Unterthiner, Mostafa Dehghani, Matthias Minderer, Georg Heigold, Sylvain Gelly, Jakob Uszkoreit, and Neil Houlsby.
\newblock An image is worth 16x16 words: Transformers for image recognition at scale.
\newblock In \emph{ICLR}, 2021.

\bibitem[Fischler and Bolles(1981)]{fischler1981random}
Martin~A Fischler and Robert~C Bolles.
\newblock Random sample consensus: a paradigm for model fitting with applications to image analysis and automated cartography.
\newblock \emph{Communications of the ACM}, 24\penalty0 (6):\penalty0 381--395, 1981.

\bibitem[Gao et~al.(2020)Gao, Saraf, Huang, and Kopf]{gao2020flow}
Chen Gao, Ayush Saraf, Jia-Bin Huang, and Johannes Kopf.
\newblock Flow-edge guided video completion.
\newblock In \emph{ECCV}, pages 713--729. Springer, 2020.

\bibitem[Geiger et~al.(2012)Geiger, Lenz, and Urtasun]{geiger2012we}
Andreas Geiger, Philip Lenz, and Raquel Urtasun.
\newblock Are we ready for autonomous driving? the kitti vision benchmark suite.
\newblock In \emph{CVPR}, pages 3354--3361. IEEE, 2012.

\bibitem[Geiger et~al.(2013)Geiger, Lenz, Stiller, and Urtasun]{kitti12}
Andreas Geiger, Philip Lenz, Christoph Stiller, and Raquel Urtasun.
\newblock Vision meets robotics: The kitti dataset.
\newblock \emph{International Journal of Robotics Research}, 32\penalty0 (11):\penalty0 1231--1237, 2013.

\bibitem[Han et~al.(2022)Han, Luo, Luo, Liu, Fan, Luo, and Liu]{han2022realflow}
Yunhui Han, Kunming Luo, Ao Luo, Jiangyu Liu, Haoqiang Fan, Guiming Luo, and Shuaicheng Liu.
\newblock Realflow: Em-based realistic optical flow dataset generation from videos.
\newblock In \emph{ECCV}, pages 288--305. Springer, 2022.

\bibitem[He et~al.(2016)He, Zhang, Ren, and Sun]{he2016identity}
Kaiming He, Xiangyu Zhang, Shaoqing Ren, and Jian Sun.
\newblock Identity mappings in deep residual networks.
\newblock In \emph{ECCV}, pages 630--645. Springer, 2016.

\bibitem[He et~al.(2023)He, Bao, Li, Grant, and Ou]{he2023accuracy}
Sheng He, Rina Bao, Jingpeng Li, P~Ellen Grant, and Yangming Ou.
\newblock Accuracy of segment-anything model (sam) in medical image segmentation tasks.
\newblock \emph{arXiv preprint arXiv:2304.09324}, 2023.

\bibitem[Horn and Schunck(1981)]{horn1981determining}
Berthold~KP Horn and Brian~G Schunck.
\newblock Determining optical flow.
\newblock \emph{Artificial Intelligence}, 17\penalty0 (1-3):\penalty0 185--203, 1981.

\bibitem[Huang et~al.(2023)Huang, Herrmann, Hur, Lu, Sargent, Stone, Yang, and Sun]{huang2023self}
Hsin-Ping Huang, Charles Herrmann, Junhwa Hur, Erika Lu, Kyle Sargent, Austin Stone, Ming-Hsuan Yang, and Deqing Sun.
\newblock Self-supervised autoflow.
\newblock In \emph{CVPR}, pages 11412--11421, 2023.

\bibitem[Huang et~al.(2022)Huang, Shi, Zhang, Wang, Cheung, Qin, Dai, and Li]{huang2022flowformer}
Zhaoyang Huang, Xiaoyu Shi, Chao Zhang, Qiang Wang, Ka~Chun Cheung, Hongwei Qin, Jifeng Dai, and Hongsheng Li.
\newblock Flowformer: A transformer architecture for optical flow.
\newblock In \emph{ECCV}, pages 668--685. Springer, 2022.

\bibitem[Hur and Roth(2016)]{hur2016joint}
Junhwa Hur and Stefan Roth.
\newblock Joint optical flow and temporally consistent semantic segmentation.
\newblock In \emph{ECCV}, pages 163--177. Springer, 2016.

\bibitem[Hur and Roth(2019)]{hur2019iterative}
Junhwa Hur and Stefan Roth.
\newblock Iterative residual refinement for joint optical flow and occlusion estimation.
\newblock In \emph{CVPR}, pages 5754--5763, 2019.

\bibitem[Im et~al.(2020)Im, Kim, and Yoon]{im2020unsupervised}
Woobin Im, Tae-Kyun Kim, and Sung-Eui Yoon.
\newblock Unsupervised learning of optical flow with deep feature similarity.
\newblock In \emph{ECCV}, pages 172--188. Springer, 2020.

\bibitem[Janai et~al.(2018)Janai, Guney, Ranjan, Black, and Geiger]{janai2018unsupervised}
Joel Janai, Fatma Guney, Anurag Ranjan, Michael Black, and Andreas Geiger.
\newblock Unsupervised learning of multi-frame optical flow with occlusions.
\newblock In \emph{ECCV}, pages 690--706, 2018.

\bibitem[Jiang et~al.(2021)Jiang, Campbell, Lu, Li, and Hartley]{jiang2021learning}
Shihao Jiang, Dylan Campbell, Yao Lu, Hongdong Li, and Richard Hartley.
\newblock Learning to estimate hidden motions with global motion aggregation.
\newblock In \emph{ICCV}, pages 9772--9781, 2021.

\bibitem[Jonschkowski et~al.(2020)Jonschkowski, Stone, Barron, Gordon, Konolige, and Angelova]{jonschkowski2020matters}
Rico Jonschkowski, Austin Stone, Jonathan~T Barron, Ariel Gordon, Kurt Konolige, and Anelia Angelova.
\newblock What matters in unsupervised optical flow.
\newblock In \emph{ECCV}, pages 557--572. Springer, 2020.

\bibitem[Jung et~al.(2023)Jung, Hui, Luo, Yang, Liu, Yoo, Ranjan, and Demandolx]{jung2023anyflow}
Hyunyoung Jung, Zhuo Hui, Lei Luo, Haitao Yang, Feng Liu, Sungjoo Yoo, Rakesh Ranjan, and Denis Demandolx.
\newblock Anyflow: Arbitrary scale optical flow with implicit neural representation.
\newblock In \emph{CVPR}, pages 5455--5465, 2023.

\bibitem[Kim et~al.(2022)Kim, Yu, Yuan, and Tomasi]{kim2022cross}
Hannah~Halin Kim, Shuzhi Yu, Shuai Yuan, and Carlo Tomasi.
\newblock Cross-attention transformer for video interpolation.
\newblock In \emph{ACCVW}, pages 320--337, 2022.

\bibitem[Kingma and Ba(2014)]{kingma2014adam}
Diederik~P Kingma and Jimmy Ba.
\newblock Adam: A method for stochastic optimization.
\newblock \emph{ICLR}, 2014.

\bibitem[Kirillov et~al.(2023)Kirillov, Mintun, Ravi, Mao, Rolland, Gustafson, Xiao, Whitehead, Berg, Lo, Dollar, and Girshick]{sam}
Alexander Kirillov, Eric Mintun, Nikhila Ravi, Hanzi Mao, Chloe Rolland, Laura Gustafson, Tete Xiao, Spencer Whitehead, Alexander~C. Berg, Wan-Yen Lo, Piotr Dollar, and Ross Girshick.
\newblock Segment anything.
\newblock In \emph{ICCV}, pages 4015--4026, 2023.

\bibitem[Krizhevsky et~al.(2012)Krizhevsky, Sutskever, and Hinton]{krizhevsky2012imagenet}
Alex Krizhevsky, Ilya Sutskever, and Geoffrey~E Hinton.
\newblock Imagenet classification with deep convolutional neural networks.
\newblock \emph{NeurIPS}, 25, 2012.

\bibitem[Liu et~al.(2020)Liu, Zhang, He, Liu, Wang, Tai, Luo, Wang, Li, and Huang]{liu2020learning}
Liang Liu, Jiangning Zhang, Ruifei He, Yong Liu, Yabiao Wang, Ying Tai, Donghao Luo, Chengjie Wang, Jilin Li, and Feiyue Huang.
\newblock Learning by analogy: Reliable supervision from transformations for unsupervised optical flow estimation.
\newblock In \emph{CVPR}, pages 6489--6498, 2020.

\bibitem[Liu et~al.(2019{\natexlab{a}})Liu, King, Lyu, and Xu]{liu2019ddflow}
Pengpeng Liu, Irwin King, Michael~R Lyu, and Jia Xu.
\newblock Ddflow: Learning optical flow with unlabeled data distillation.
\newblock In \emph{AAAI}, pages 8770--8777, 2019{\natexlab{a}}.

\bibitem[Liu et~al.(2019{\natexlab{b}})Liu, Lyu, King, and Xu]{liu2019selflow}
Pengpeng Liu, Michael Lyu, Irwin King, and Jia Xu.
\newblock Selflow: Self-supervised learning of optical flow.
\newblock In \emph{CVPR}, pages 4571--4580, 2019{\natexlab{b}}.

\bibitem[Liu et~al.(2021)Liu, Luo, Ye, Wang, Wang, and Zeng]{liu2021oiflow}
Shuaicheng Liu, Kunming Luo, Nianjin Ye, Chuan Wang, Jue Wang, and Bing Zeng.
\newblock Oiflow: Occlusion-inpainting optical flow estimation by unsupervised learning.
\newblock \emph{IEEE TIP}, 30:\penalty0 6420--6433, 2021.

\bibitem[Lucas and Kanade(1981)]{lucas1981iterative}
Bruce~D Lucas and Takeo Kanade.
\newblock An iterative image registration technique with an application to stereo vision.
\newblock In \emph{IJCAI}, pages 674--679, 1981.

\bibitem[Luo et~al.(2022)Luo, Yang, Luo, Li, Fan, and Liu]{luo2022learning}
Ao Luo, Fan Yang, Kunming Luo, Xin Li, Haoqiang Fan, and Shuaicheng Liu.
\newblock Learning optical flow with adaptive graph reasoning.
\newblock In \emph{AAAI}, pages 1890--1898, 2022.

\bibitem[Luo et~al.(2021)Luo, Wang, Liu, Fan, Wang, and Sun]{luo2021upflow}
Kunming Luo, Chuan Wang, Shuaicheng Liu, Haoqiang Fan, Jue Wang, and Jian Sun.
\newblock Upflow: Upsampling pyramid for unsupervised optical flow learning.
\newblock In \emph{CVPR}, pages 1045--1054, 2021.

\bibitem[Ma and Wang(2023)]{ma2023segment}
Jun Ma and Bo Wang.
\newblock Segment anything in medical images.
\newblock \emph{arXiv preprint arXiv:2304.12306}, 2023.

\bibitem[Mazurowski et~al.(2023)Mazurowski, Dong, Gu, Yang, Konz, and Zhang]{mazurowski2023segment}
Maciej~A Mazurowski, Haoyu Dong, Hanxue Gu, Jichen Yang, Nicholas Konz, and Yixin Zhang.
\newblock Segment anything model for medical image analysis: an experimental study.
\newblock \emph{Medical Image Analysis}, 89:\penalty0 102918, 2023.

\bibitem[Meister et~al.(2018)Meister, Hur, and Roth]{meister2018unflow}
Simon Meister, Junhwa Hur, and Stefan Roth.
\newblock Unflow: Unsupervised learning of optical flow with a bidirectional census loss.
\newblock In \emph{AAAI}, 2018.

\bibitem[Menze and Geiger(2015)]{kitti15}
Moritz Menze and Andreas Geiger.
\newblock Object scene flow for autonomous vehicles.
\newblock In \emph{CVPR}, 2015.

\bibitem[Neubeck and Van~Gool(2006)]{neubeck2006efficient}
Alexander Neubeck and Luc Van~Gool.
\newblock Efficient non-maximum suppression.
\newblock In \emph{International Conference on Pattern Recognition}, pages 850--855. IEEE, 2006.

\bibitem[Paszke et~al.(2019)Paszke, Gross, Massa, Lerer, Bradbury, Chanan, Killeen, Lin, Gimelshein, Antiga, Desmaison, Kopf, Yang, DeVito, Raison, Tejani, Chilamkurthy, Steiner, Fang, Bai, and Chintala]{NEURIPS2019_9015}
Adam Paszke, Sam Gross, Francisco Massa, Adam Lerer, James Bradbury, Gregory Chanan, Trevor Killeen, Zeming Lin, Natalia Gimelshein, Luca Antiga, Alban Desmaison, Andreas Kopf, Edward Yang, Zachary DeVito, Martin Raison, Alykhan Tejani, Sasank Chilamkurthy, Benoit Steiner, Lu Fang, Junjie Bai, and Soumith Chintala.
\newblock Pytorch: An imperative style, high-performance deep learning library.
\newblock In \emph{NeurIPS}, pages 8024--8035. Curran Associates, Inc., 2019.

\bibitem[Qi et~al.(2017)Qi, Su, Mo, and Guibas]{qi2017pointnet}
Charles~R Qi, Hao Su, Kaichun Mo, and Leonidas~J Guibas.
\newblock Pointnet: Deep learning on point sets for 3d classification and segmentation.
\newblock In \emph{CVPR}, pages 652--660, 2017.

\bibitem[Raji{\v{c}} et~al.(2023)Raji{\v{c}}, Ke, Tai, Tang, Danelljan, and Yu]{rajivc2023segment}
Frano Raji{\v{c}}, Lei Ke, Yu-Wing Tai, Chi-Keung Tang, Martin Danelljan, and Fisher Yu.
\newblock Segment anything meets point tracking.
\newblock \emph{arXiv preprint arXiv:2307.01197}, 2023.

\bibitem[Ren et~al.(2017)Ren, Yan, Ni, Liu, Yang, and Zha]{ren2017unsupervised}
Zhe Ren, Junchi Yan, Bingbing Ni, Bin Liu, Xiaokang Yang, and Hongyuan Zha.
\newblock Unsupervised deep learning for optical flow estimation.
\newblock In \emph{AAAI}, 2017.

\bibitem[Sevilla-Lara et~al.(2016)Sevilla-Lara, Sun, Jampani, and Black]{sevilla2016optical}
Laura Sevilla-Lara, Deqing Sun, Varun Jampani, and Michael~J Black.
\newblock Optical flow with semantic segmentation and localized layers.
\newblock In \emph{CVPR}, pages 3889--3898, 2016.

\bibitem[Shen et~al.(2023)Shen, Yang, and Wang]{shen2023anything}
Qiuhong Shen, Xingyi Yang, and Xinchao Wang.
\newblock Anything-3d: Towards single-view anything reconstruction in the wild.
\newblock \emph{arXiv preprint arXiv:2304.10261}, 2023.

\bibitem[Shi et~al.(2023)Shi, Huang, Li, Zhang, Cheung, See, Qin, Dai, and Li]{shi2023flowformer++}
Xiaoyu Shi, Zhaoyang Huang, Dasong Li, Manyuan Zhang, Ka~Chun Cheung, Simon See, Hongwei Qin, Jifeng Dai, and Hongsheng Li.
\newblock Flowformer++: Masked cost volume autoencoding for pretraining optical flow estimation.
\newblock In \emph{CVPR}, pages 1599--1610, 2023.

\bibitem[Smith and Topin(2019)]{smith2019super}
Leslie~N Smith and Nicholay Topin.
\newblock Super-convergence: Very fast training of neural networks using large learning rates.
\newblock In \emph{Artificial Intelligence and Machine Learning for Multi-domain Operations Applications}, pages 369--386. SPIE, 2019.

\bibitem[Stone et~al.(2021)Stone, Maurer, Ayvaci, Angelova, and Jonschkowski]{stone2021smurf}
Austin Stone, Daniel Maurer, Alper Ayvaci, Anelia Angelova, and Rico Jonschkowski.
\newblock Smurf: Self-teaching multi-frame unsupervised raft with full-image warping.
\newblock In \emph{CVPR}, pages 3887--3896, 2021.

\bibitem[Sun et~al.(2018)Sun, Yang, Liu, and Kautz]{sun2018pwc}
Deqing Sun, Xiaodong Yang, Ming-Yu Liu, and Jan Kautz.
\newblock Pwc-net: Cnns for optical flow using pyramid, warping, and cost volume.
\newblock In \emph{CVPR}, pages 8934--8943, 2018.

\bibitem[Sun et~al.(2019)Sun, Yang, Liu, and Kautz]{sun2019models}
Deqing Sun, Xiaodong Yang, Ming-Yu Liu, and Jan Kautz.
\newblock Models matter, so does training: An empirical study of cnns for optical flow estimation.
\newblock \emph{IEEE TPAMI}, 42\penalty0 (6):\penalty0 1408--1423, 2019.

\bibitem[Sun et~al.(2021)Sun, Vlasic, Herrmann, Jampani, Krainin, Chang, Zabih, Freeman, and Liu]{sun2021autoflow}
Deqing Sun, Daniel Vlasic, Charles Herrmann, Varun Jampani, Michael Krainin, Huiwen Chang, Ramin Zabih, William~T Freeman, and Ce Liu.
\newblock Autoflow: Learning a better training set for optical flow.
\newblock In \emph{CVPR}, pages 10093--10102, 2021.

\bibitem[Teed and Deng(2020)]{teed2020raft}
Zachary Teed and Jia Deng.
\newblock Raft: Recurrent all-pairs field transforms for optical flow.
\newblock In \emph{ECCV}, pages 402--419. Springer, 2020.

\bibitem[Wang et~al.(2019)Wang, Zhu, Liu, Ye, Li, and Zhang]{wang2019semflow}
Xianshun Wang, Dongchen Zhu, Yanqing Liu, Xiaoqing Ye, Jiamao Li, and Xiaolin Zhang.
\newblock Semflow: Semantic-driven interpolation for large displacement optical flow.
\newblock \emph{IEEE Access}, 7:\penalty0 51589--51597, 2019.

\bibitem[Wang et~al.(2018)Wang, Yang, Yang, Zhao, Wang, and Xu]{wang2018occlusion}
Yang Wang, Yi Yang, Zhenheng Yang, Liang Zhao, Peng Wang, and Wei Xu.
\newblock Occlusion aware unsupervised learning of optical flow.
\newblock In \emph{CVPR}, pages 4884--4893, 2018.

\bibitem[Wang et~al.(2004)Wang, Bovik, Sheikh, and Simoncelli]{wang2004image}
Zhou Wang, Alan~C Bovik, Hamid~R Sheikh, and Eero~P Simoncelli.
\newblock Image quality assessment: from error visibility to structural similarity.
\newblock \emph{IEEE TIP}, 13\penalty0 (4):\penalty0 600--612, 2004.

\bibitem[Wu et~al.(2023)Wu, Fu, Fang, Liu, Wang, Xu, Jin, and Arbel]{wu2023medical}
Junde Wu, Rao Fu, Huihui Fang, Yuanpei Liu, Zhaowei Wang, Yanwu Xu, Yueming Jin, and Tal Arbel.
\newblock Medical sam adapter: Adapting segment anything model for medical image segmentation.
\newblock \emph{arXiv preprint arXiv:2304.12620}, 2023.

\bibitem[Wulff et~al.(2017)Wulff, Sevilla-Lara, and Black]{wulff2017optical}
Jonas Wulff, Laura Sevilla-Lara, and Michael~J Black.
\newblock Optical flow in mostly rigid scenes.
\newblock In \emph{CVPR}, pages 4671--4680, 2017.

\bibitem[Yang and Yezzi(2022)]{yang2022decomposing}
Huizong Yang and Anthony Yezzi.
\newblock Decomposing the tangent of occluding boundaries according to curvatures and torsions.
\newblock In \emph{European Conference on Computer Vision}, pages 123--138. Springer, 2022.

\bibitem[Yang et~al.(2023)Yang, Gao, Li, Gao, Wang, and Zheng]{yang2023track}
Jinyu Yang, Mingqi Gao, Zhe Li, Shang Gao, Fangjing Wang, and Feng Zheng.
\newblock Track anything: Segment anything meets videos.
\newblock \emph{arXiv preprint arXiv:2304.11968}, 2023.

\bibitem[Ye et~al.(2022)Ye, Li, Tucker, Kanazawa, and Snavely]{ye2022deformable}
Vickie Ye, Zhengqi Li, Richard Tucker, Angjoo Kanazawa, and Noah Snavely.
\newblock Deformable sprites for unsupervised video decomposition.
\newblock In \emph{CVPR}, pages 2657--2666, 2022.

\bibitem[Yu et~al.(2016)Yu, Harley, and Derpanis]{yu2016back}
Jason~J Yu, Adam~W Harley, and Konstantinos~G Derpanis.
\newblock Back to basics: Unsupervised learning of optical flow via brightness constancy and motion smoothness.
\newblock In \emph{ECCVW}, pages 3--10. Springer, 2016.

\bibitem[Yu et~al.(2022)Yu, Kim, Yuan, and Tomasi]{yu2022unsupervised}
Shuzhi Yu, Hannah~H Kim, Shuai Yuan, and Carlo Tomasi.
\newblock Unsupervised flow refinement near motion boundaries.
\newblock In \emph{BMVC}. {BMVA} Press, 2022.

\bibitem[Yuan(2023)]{yuan2023assisting}
Shuai Yuan.
\newblock \emph{Assisting Unsupervised Optical Flow Estimation With External Information}.
\newblock PhD thesis, Duke University, 2023.

\bibitem[Yuan and Tomasi(2023)]{yuan2023ufd}
Shuai Yuan and Carlo Tomasi.
\newblock Ufd-prime: Unsupervised joint learning of optical flow and stereo depth through pixel-level rigid motion estimation.
\newblock \emph{arXiv preprint arXiv:2310.04712}, 2023.

\bibitem[Yuan et~al.(2022)Yuan, Sun, Kim, Yu, and Tomasi]{yuan2022optical}
Shuai Yuan, Xian Sun, Hannah Kim, Shuzhi Yu, and Carlo Tomasi.
\newblock Optical flow training under limited label budget via active learning.
\newblock In \emph{ECCV}, pages 410--427. Springer Nature Switzerland, 2022.

\bibitem[Yuan et~al.(2023)Yuan, Yu, Kim, and Tomasi]{yuan2023semarflow}
Shuai Yuan, Shuzhi Yu, Hannah Kim, and Carlo Tomasi.
\newblock Semarflow: Injecting semantics into unsupervised optical flow estimation for autonomous driving.
\newblock In \emph{ICCV}, pages 9566--9577, 2023.

\bibitem[Zach et~al.(2007)Zach, Pock, and Bischof]{zach2007duality}
Christopher Zach, Thomas Pock, and Horst Bischof.
\newblock A duality based approach for realtime tv-l 1 optical flow.
\newblock In \emph{Pattern Recognition}, pages 214--223. Springer, 2007.

\bibitem[Zhang et~al.(2021)Zhang, Woodford, Prisacariu, and Torr]{zhang2021separable}
Feihu Zhang, Oliver~J Woodford, Victor~Adrian Prisacariu, and Philip~HS Torr.
\newblock Separable flow: Learning motion cost volumes for optical flow estimation.
\newblock In \emph{ICCV}, pages 10807--10817, 2021.

\bibitem[Zhang et~al.(2023)Zhang, Wei, Zhang, Dai, and Zhu]{zhang2023uvosam}
Zhenghao Zhang, Zhichao Wei, Shengfan Zhang, Zuozhuo Dai, and Siyu Zhu.
\newblock Uvosam: A mask-free paradigm for unsupervised video object segmentation via segment anything model.
\newblock \emph{arXiv preprint arXiv:2305.12659}, 2023.

\bibitem[Zhou et~al.(2023)Zhou, He, Tan, and Yan]{zhou2023samflow}
Shili Zhou, Ruian He, Weimin Tan, and Bo Yan.
\newblock Samflow: Eliminating any fragmentation in optical flow with segment anything model.
\newblock \emph{arXiv preprint arXiv:2307.16586}, 2023.

\bibitem[Zhu et~al.(2019)Zhu, Sapra, Reda, Shih, Newsam, Tao, and Catanzaro]{zhu2019improving}
Yi Zhu, Karan Sapra, Fitsum~A Reda, Kevin~J Shih, Shawn Newsam, Andrew Tao, and Bryan Catanzaro.
\newblock Improving semantic segmentation via video propagation and label relaxation.
\newblock In \emph{CVPR}, pages 8856--8865, 2019.

\end{thebibliography}
}

\clearpage
\newpage
\doparttoc % Tell to minitoc to generate a toc for the parts
\faketableofcontents % Run a fake tableofcontents command for the partocs

\appendix
\addcontentsline{toc}{section}{Appendix} % Add the appendix text to the document TOC
\part{Appendix} % Start the appendix part
\parttoc % Insert the appendix TOC
\section{Method details}

\subsection{Network structures}

\paragraph{Why ARFlow as backbone?}

We choose ARFlow~\cite{liu2020learning} as our backbone in light of the following considerations.
\begin{itemize}
    \item Our main goal is to investigate how SAM~\cite{sam} can help with unsupervised optical flow instead of pushing the best possible performances by all means. Therefore, adopting a simple yet effective model, such as ARFlow, works better for our research.
    \item ARFlow is especially light-weight and easy to train. Previous related work, SemARFlow~\cite{yuan2023semarflow} also adopts the ARFlow backbone, so we follow them to  borrow similar ideas from SemARFlow as well.
    \item Previous research has shown that ARFlow can be easily adapted to have close to UPFlow performances while maintaining simplicity~\cite{yuan2023semarflow,yuan2023ufd}. We follow those suggestions and build our own baseline model, evaluated in the experiments section.
\end{itemize}

\paragraph{Why is there no comparisons with SMURF?} 

Admittedly, SMURF~\cite{stone2021smurf} has achieved outstanding performances on unsupervised optical flow estimation. However, we do not compare with them in the experiments section for the following reasons.
\begin{itemize}
    \item As mentioned above, our goal is to see how SAM~\cite{sam} can help with unsupervised optical flow instead of pushing the best possible performances by all means. We compare our models with the baseline model that does not apply SAM to show how SAM is effective, while other previous methods are shown as references to help understand in absolute terms how our adapted version performs.
    \item SMURF uses a larger architecture RAFT~\cite{teed2020raft}, and arguably, its success highly relies on its technical designs such as full-image warping, multi-frame self-supervision (which requires training a tiny model for each training sample separately), as well as its extensive data augmentations. We could definitely keep these technical designs in our model as well to enhance performances. However, they add great complexities to our network and may greatly increase the computational costs of our experiments. Therefore, we choose to stay simple and focus more on how to inject SAM information effectively. 
\end{itemize}

\begin{figure*}
    \centering
    \includegraphics[width=0.66\linewidth]{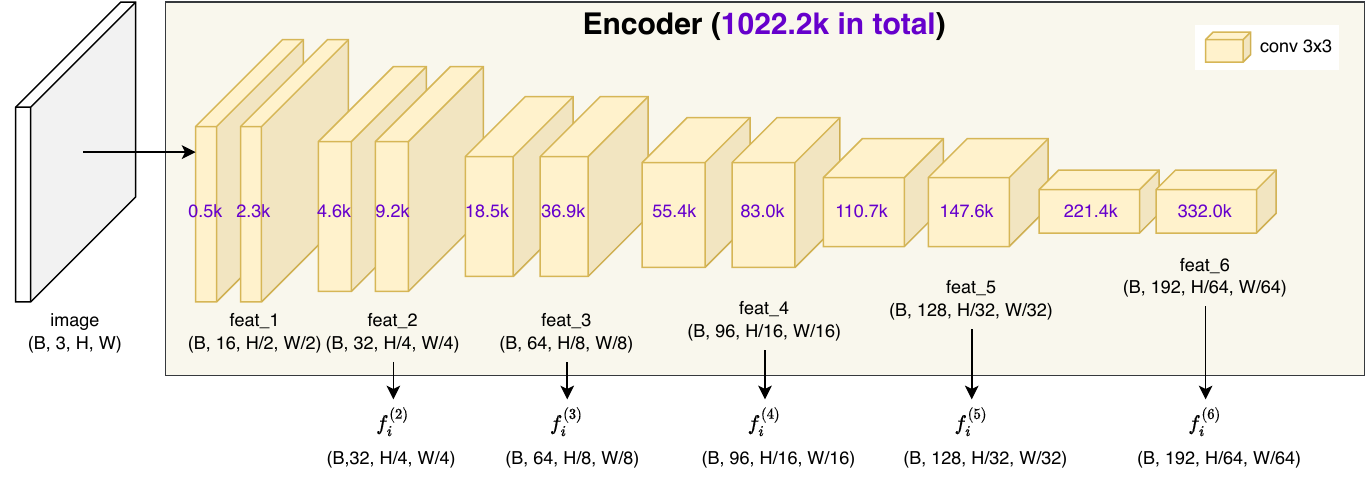}
    \caption{Detailed encoder structure (figure adapted from ~\cite{yuan2023ufd}); numbers in purple refer to the parameter sizes of each module.}
    \label{fig:enc_app}
\end{figure*}

\begin{figure*}
    \centering
    \includegraphics[width=\linewidth]{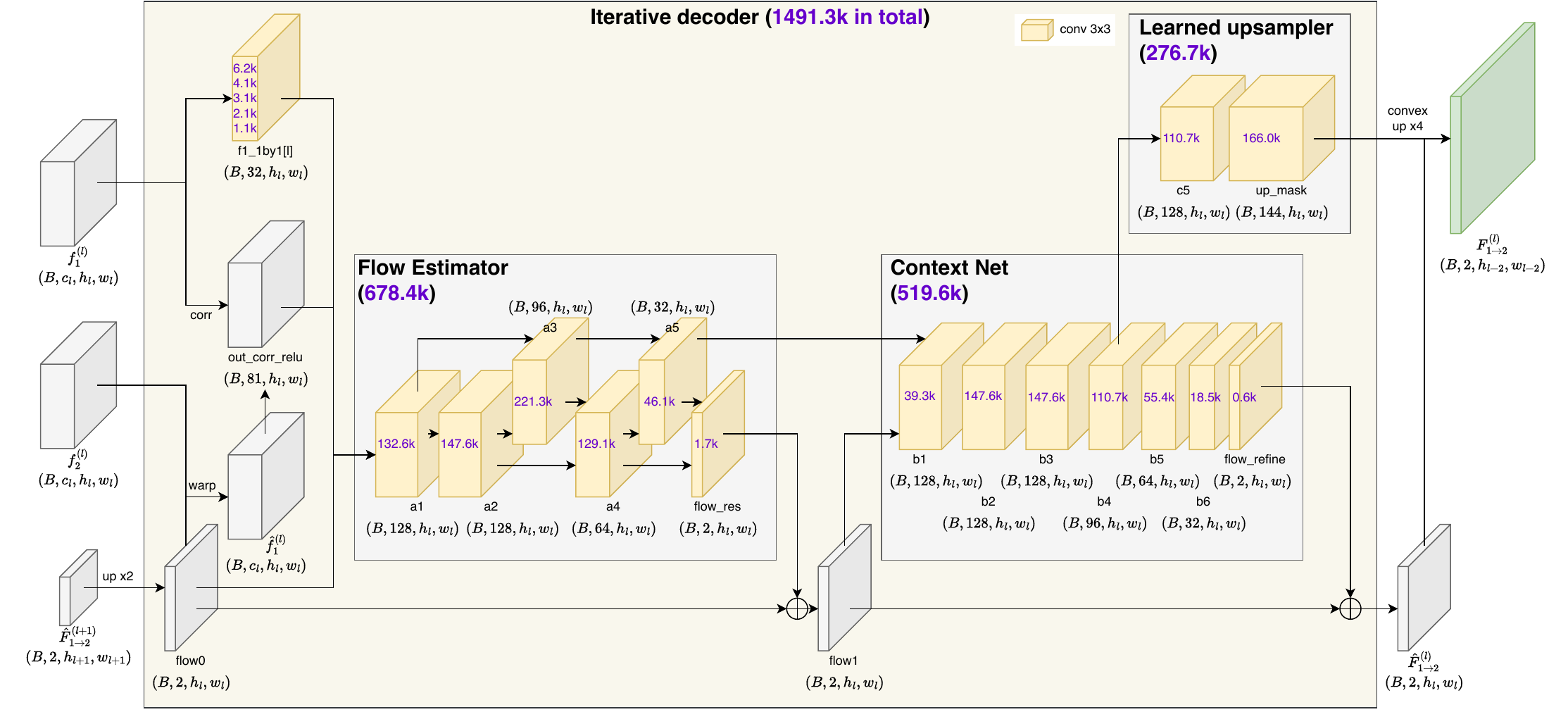}
    \caption{Detailed decoder structure (figure adapted from ~\cite{yuan2023ufd}); numbers in purple refer to the parameter sizes. The mask feature modules are not shown in the figure for conciseness. We use the same warping and correlation computation for mask feature and image features.}
    \label{fig:dec_app}
\end{figure*}

\paragraph{Detailed structures}
Our detailed network structures are shown in \cref{fig:enc_app,fig:dec_app}.

\subsection{Decoder computation}

As shown in Fig. 2b in the main paper. For each iteration on level $l$, the decoder takes in image features $f_1^{(l)}$, $f_2^{(l)}$ and the flow estimate from the previous level $\hat F_{1\to 2}^{(l+1)}$ (and also masks $M_1$, $M_2$ if the mask feature module is turned on), and outputs the refined flow estimate at the current level  $\hat F_{1\to 2}^{(l)}$.

A 1-by-1 convolution layer is first applied to $f_1^{(l)}$ to unify the feature channel sizes of different levels to be the same number 32. This module enables us to reuse the same following modules on all levels. We first upsample $\hat F_{1\to 2}^{(l+1)}$ by 2 times to the same resolution $\tilde F_{1\to 2}^{(l)}$ as the features at this level through a simple bilinear interpolation. The upsampled flow is then used to warp $f_2^{(l)}$ as
\begin{equation}
    \hat f_1^{(l)} (\bm p) = f_2^{(l)} (\bm p+\tilde F_{1\to 2}^{(l)}),\quad \forall\bm p
\end{equation}
which can be implemented through a grid sampling process.

We then compute the correlation between $f_1^{(l)}$ and $\hat f_1^{(l)}$ through a $9\times 9$ neighborhood window, yielding a flattened 81-channel correlation output. The correlation is then concatenated with the upsampled flow $\tilde F_{1\to 2}^{(l)}$ and the 1-by-1 convolutional features as the input to the flow estimator network. Note that if the mask feature module is turned on, we also do warping and correlation to the mask feature $g_t^{(l)}$ in the same way.

The flow estimator estimates a flow residual added to the current estimate $\tilde F_{1\to 2}^{(l)}$. Subsequently, a context net is also applied similarly to obtain the refined flow of this level $\hat F_{1\to 2}^{(l)}$. The learned upsampler, adapted from the one in RAFT~\cite{teed2020raft}, outputs the parameters for the convex upsampling of $\hat F_{1\to 2}^{(l)}$, yielding the upsampled final output $F_{1\to 2}^{(l)}$.

\subsection{Semantic augmentation}

\paragraph{Heuristic} Our heuristic for choosing key objects is that a key object may have many object parts that could be also detected by SAM, so the key object masks may overlap with many other object masks.

\begin{figure}[htb]
    \centering
    \includegraphics[width=\linewidth]{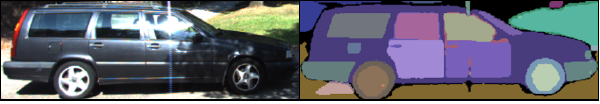}
    \caption{Example of the SAM masks computed for a car patch}
    \label{fig:aug_demo_app}
\end{figure}

One example is shown in \cref{fig:aug_demo_app}. Not only has the whole car object been detected by SAM, but also its components such as front and rear wheels, car doors and windows, lights, bumpers, and even the gas cap. As a result, the mask of the whole car object overlaps with all those component masks, whereas each component mask only overlaps with the car mask. Thus, the car object will be selected due to its high degree of mask overlap. Empirically, our heuristic works on all car objects pretty well, which are indeed key objects in autonomous driving.

\paragraph{Key object selection}
We discuss more details on the process that we select key objects from the SAM masks. Suppose for the input image $I$, a number of $n$ masks are detected by SAM, constituting masks $M\in\{0, 1\}^{n\times H\times W}$. Denote $M(k)\in\{0, 1\}^{H\times W}$ as the $k$-th mask, and $M(k, i, j)=1$ means the pixel $(i, j)$ is on the $k$-th object. For each mask $M(k)$, we examine the following.
\begin{itemize}
    \item We first filter masks at a certain dimension. Suppose the bounding box of $M(k)$ has dimension $h\times w$, we only accept masks with $50\leq h\leq 200$, $50\leq w\leq 400$. We avoid too large masks because they may not fit in the new sample well in our augmentation. We avoid too small masks as they make little difference in the augmentation.
    \item We drop the mask if the area of the mask is smaller than 50\% of its bounding box area, \ie $\sum_{i, j}M(k, i, j) < 50\%\cdot hw$. This rule is used to exclude those severely occluded objects.
    \item We accept the mask if it overlaps with at least 5 other masks. The number of overlaps can be counted efficiently though matrix computation of $M$.
\end{itemize}

\paragraph{Training steps}
During key object selection, we save the selected masks for each training sample on the disk before starting to train, so this step adds little time or memory consumption during training. For each training sample, we load three key objects from the object cache for augmentation. For more details about semantic augmentation, we refer readers to the original paper of SemARFlow~\cite{yuan2023semarflow} and ARFlow~\cite{liu2020learning}.

\subsection{Homography smoothness loss}

\paragraph{Selecting object regions of interest}

Before selecting object regions, we first transform our raw SAM masks $M_t$ to its full segmentation representation as described in Sec. 3.5 in the main paper. This makes sure that we do not refine the same pixel multiple times. Also, the segmentation is usually smaller pieces of objects, where homography is more likely to work well.

We estimate the occlusion region using forward-backward consistency check~\cite{meister2018unflow}, as we did when computing photometric loss. The estimated occlusion region is a good cue of where the current flow estimate is less reliable. Then, we count the number of occlusion pixels for each segmentation in the full segmentation representation and pick the top six as candidates. Empirically, we find that six segmentation regions can generally cover most of the occluded pixels. Although including more candidates can improve performance, the improvement comes at a larger computational cost.

\paragraph{Refining each selected regions}
For each of the candidate regions selected above, we first find all the correspondences in that region from flow. We define the reliable flow as those non-occluded parts estimated above. We only proceed if the reliable flow part accounts for at least 20\% of the whole region.

Using the reliable flow correspondences, we estimate homography using RANSAC~\cite{fischler1981random} and compute the inlier percantage of this computation based on reprojection error. We only accept the homography if inlier percentage is at least 50\%.

Consequently, using the accepted homography, we refine the correspondences of every pixel in the object region and generate refined flow.

\begin{figure}
    \centering
    \includegraphics[width=\linewidth]{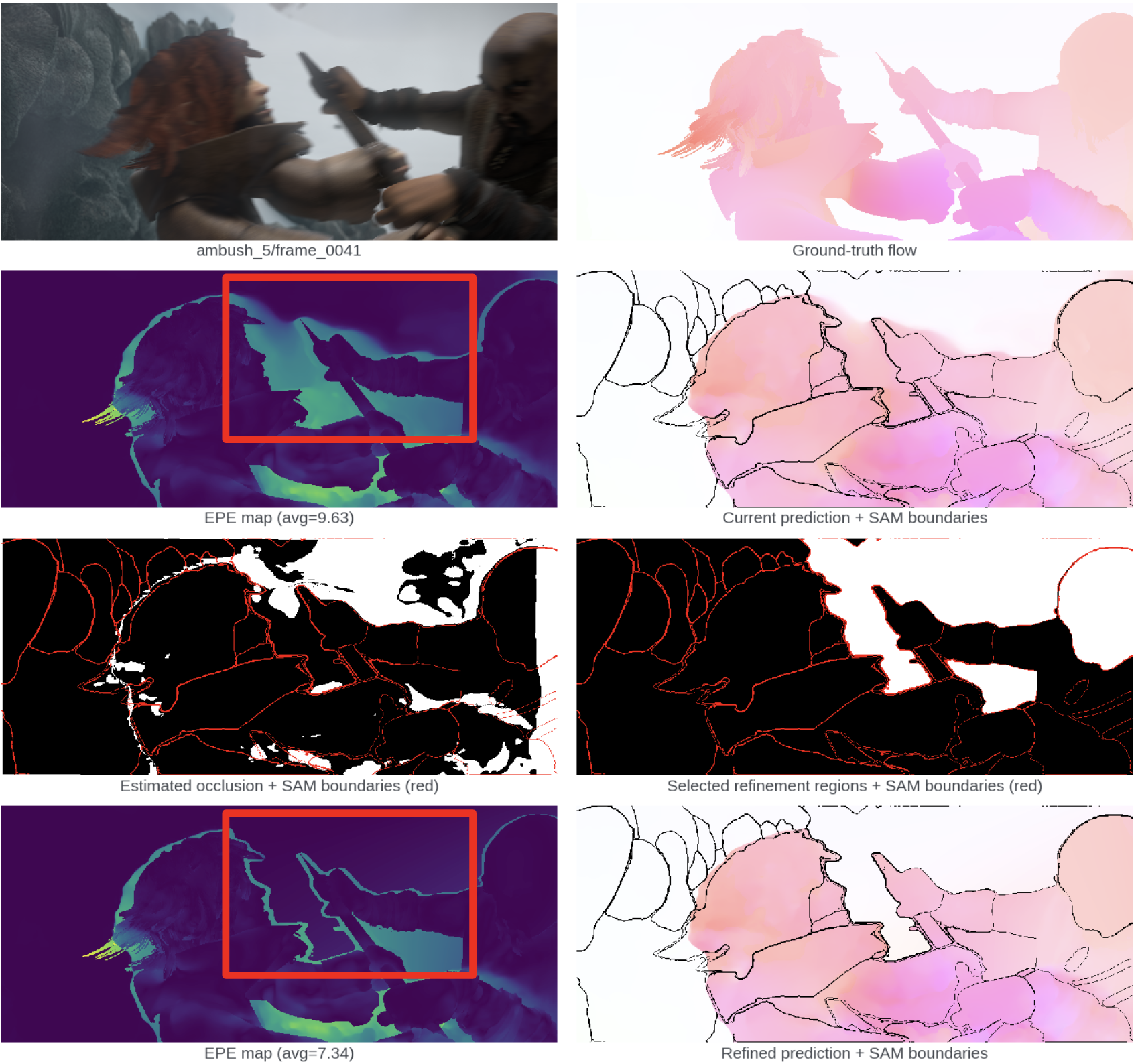}
    \caption{An example for homography refinement (Sintel)}
    \label{fig:hg_app_s1}
\end{figure}

\begin{figure}
    \centering
    \includegraphics[width=\linewidth]{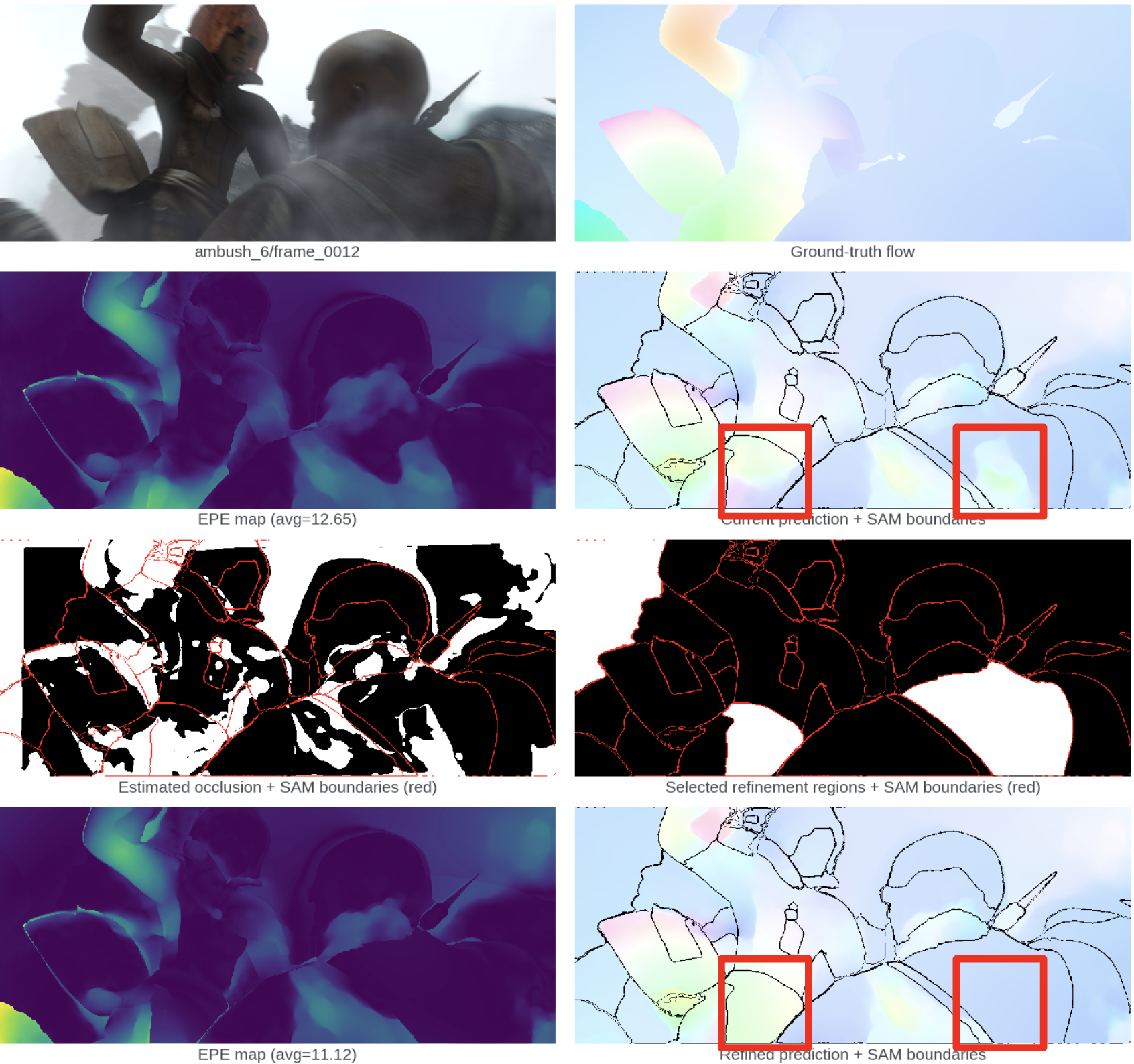}
    \caption{Another example for homography refinement (Sintel)}
    \label{fig:hg_app_s2}
\end{figure}

\begin{figure}
    \centering
    \includegraphics[width=\linewidth]{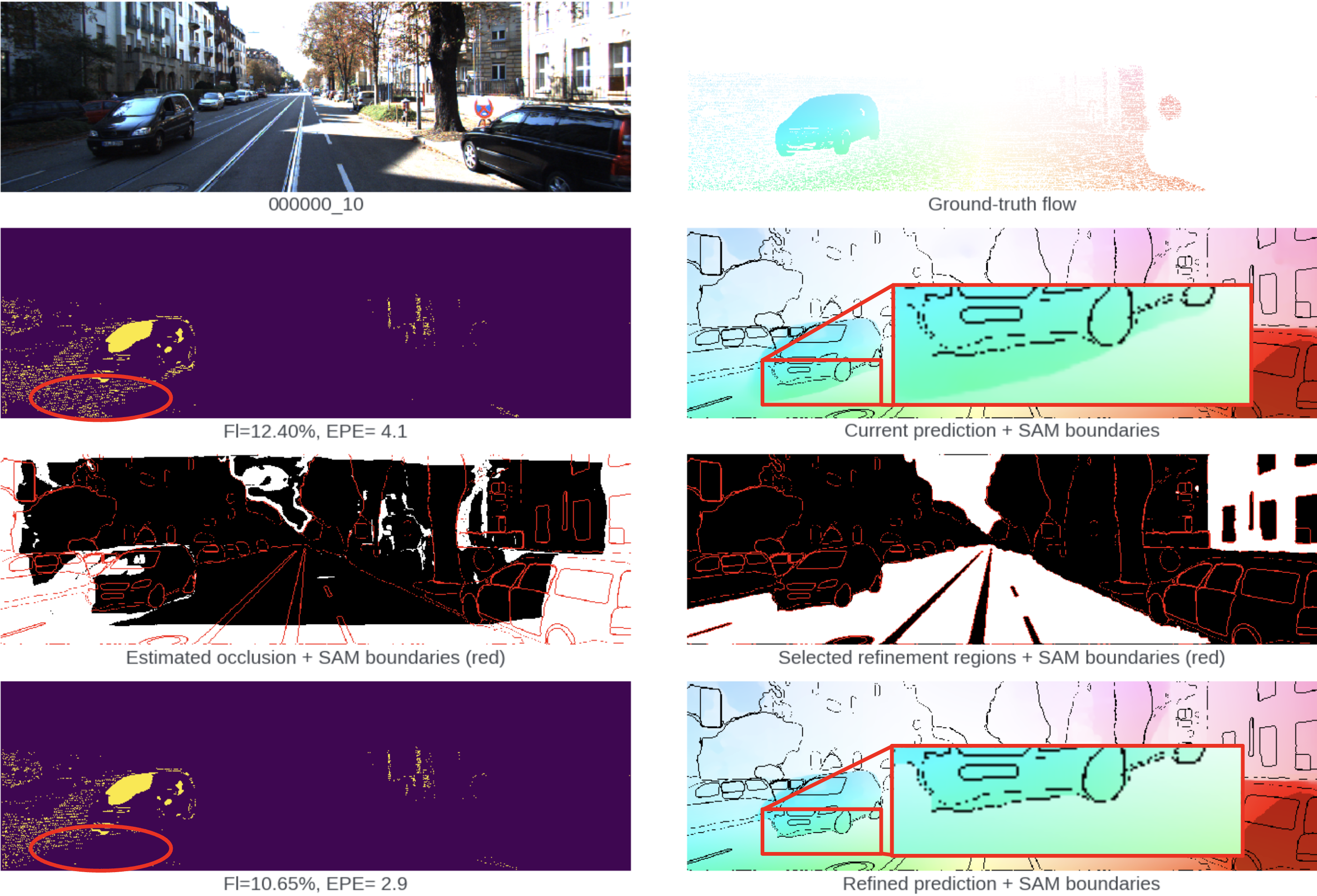}
    \caption{An example for homography refinement (KITTI)}
    \label{fig:hg_app_k}
\end{figure}

\paragraph{Examples}
Some examples of the refined flow using homography are shown in \cref{fig:hg_app_s1,fig:hg_app_s2,fig:hg_app_k}.

\paragraph{Alleviating limitations of homography}
Admittedly, homography is not precise for all objects. It mostly works well on planar, rigid regions where no deformation occurs. In our method, we alleviate his issue through the following rules.

\begin{itemize}
    \item We use full segmentation regions mentioned above, which generally refer to small object parts instead of the large object. Although the whole object may has very complex motions, its small parts are more likely to follow homography constraints.
    \item We introduce many rigorous accept/reject criteria mentioned above. If there is any sign that the homography relationship does not hold for the specific region, we stop using it. Only the most reliable homogrphies are used in refinement.
    \item We apply homography in the smoothness loss definition instead of direct post-processing. This allows our network to leverage between homography and other motion cues such as photometric constraints. Thus, a poor homography (if any) may not have large impacts if other signs/losses do not agree.
\end{itemize}

In addition, to better resolve this issue, it may be better if we could also obtain the semantic class of each object mask or if we could use text prompt to find masks, which may be updated in the later SAM versions.

\subsection{Mask feature and correlation}

Below are certain points that we need to take care when designing the mask feature module.

\begin{itemize}
    \item The raw SAM masks are discrete and arbitrary (see Sec. 3.1 for explanations). The number of masks in each sample is not fixed. The masks can have different shapes and sizes. The masks can overlap or leave holes (parts that do not belong to any masks) in the frame. Therefore, we first standardize the mask representation using a full segmentation representation described in Sec. 3.5.
    \item Our mask feature module should be independent of the order of masks, \ie our mask feature should be invariant against any permutation of masks. The mask/object IDs in the masks can be permuted without changing the segmentation map. Therefore, in our proposed module, we extract features for each mask separately regardless of its order.
    \item When we extract mask feature for each mask, the shape and size may vary, so our designed module should be well-defined for inputs of any size. This is why traditional convolutional layers may not work directly. Inspired by PointNet~\cite{qi2017pointnet}, for which the input size can also vary, we adopt operators like averaging or min/max to aggregate features of variable sizes. We apply max pooling in favor of average pooling because it adds non-linearity to the network and is often used in image classification networks. Apart from that, we need a new feature space where the max operation works. That is why we add the 1-by-1 convolutional layer at the front.
    \item The pooled feature is the same for every pixel in the same mask. This may cause numerical issues in optmization. Therefore, we concatenate and add another 1-by-1 convolutional layer to make sure the output mask feature is not exactly the same everywhere in the same mask.
\end{itemize}

\subsection{Additional explanation on Fig. 4f in the paper}
\paragraph{Why is there no curve for our proposed smoothness loss as a comparison?}  The way how our homography loss works is different. For traditional loss, since its gradients only concentrate around the flow boundary (Fig. 4d), the gradients push boundaries towards the optimal solution step by step, so we draw this landscape in Fig. 4f as if the flow boundary is moving. However, for our homography loss, the gradients apply on the whole region directly and instantly (Fig. 4e), so they are not just pushing the boundaries. Therefore, the same analysis in Fig. 4f may not apply.

\section{Result details}

\subsection{Benchmark test screenshots}

\begin{figure}[htb]
  \centering
  \begin{subfigure}{0.4\linewidth}
    \includegraphics[width=\linewidth]{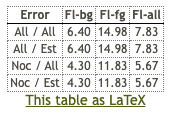}
    \caption{KITTI-2015~\cite{kitti15} results}
    \label{fig:kitti2015_benchmark}
  \end{subfigure}
  \hfill
  \begin{subfigure}{0.58\linewidth}
    \includegraphics[width=\linewidth]{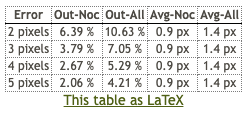}
    \caption{KITTI-2012~\cite{kitti12} test results}
    \label{fig:kitti2012_benchmark}
  \end{subfigure}
\caption{Detailed test results of our final model on KITTI}
\label{fig:kitti_benchmark}
\end{figure}

\begin{figure*}[htb]
    \centering
    \includegraphics[width=\linewidth]{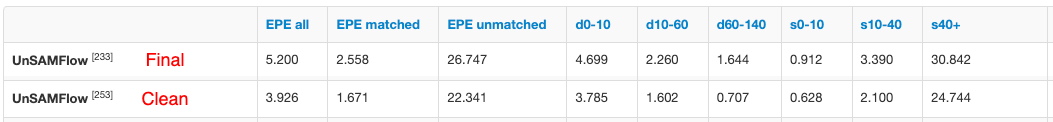}
    \caption{Detailed test results of our final model on Sintel~\cite{sintel}}
    \label{fig:sintel_benchmark}
\end{figure*}

In \cref{fig:kitti_benchmark,fig:sintel_benchmark}, we show the benchmark test screenshots of our final model on KITTI and Sintel with more detailed evaluation metrics.

\subsection{More qualitative examples}

\begin{figure*}[tb]
    \centering
    \includegraphics[width=\linewidth]{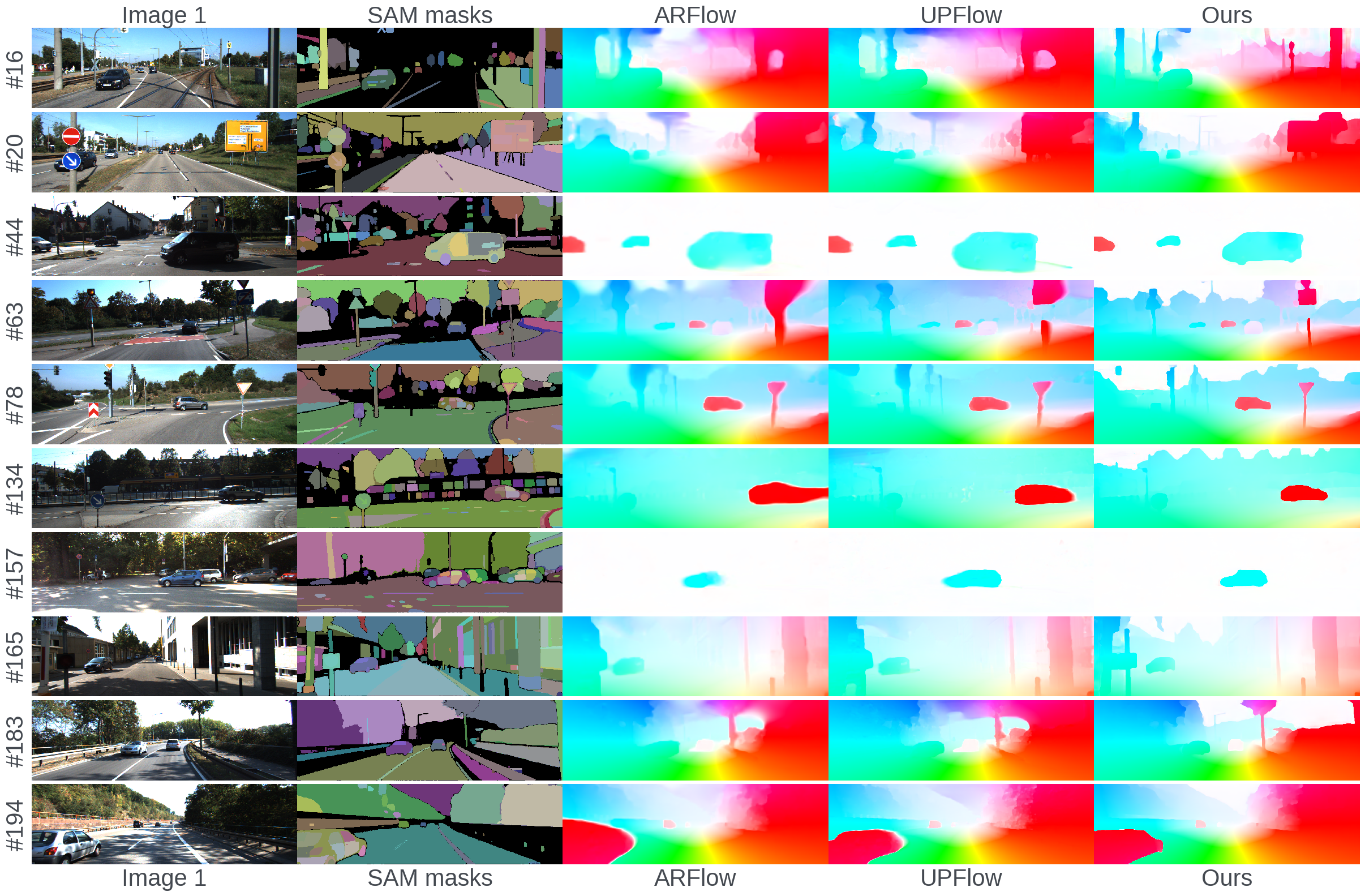}
    \caption{More qualitative results on KITTI-2015 test set~\cite{kitti15}}
    \label{fig:kitti_qual_appendix}
\end{figure*}

\begin{figure*}[tb]
    \centering
    \includegraphics[width=\linewidth]{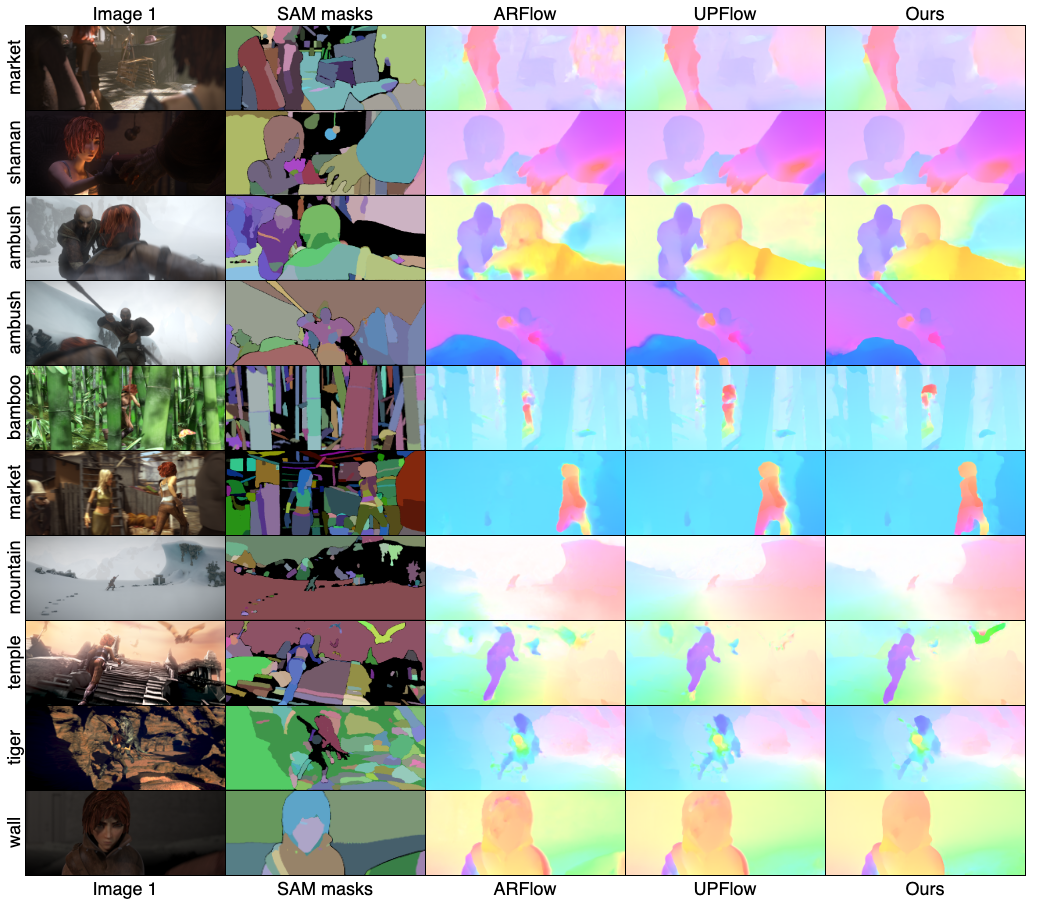}
    \caption{More qualitative results on Sintel test set~\cite{sintel}}
    \label{fig:sintel_qual_appendix}
\end{figure*}

We show more qualitative examples from the test set of KITTI-2015 (\cref{fig:kitti_qual_appendix}) and Sintel (\cref{fig:sintel_qual_appendix}).

\subsection{Experiment timing}

\paragraph{Inference} As shown in Sec 4.7 in the main paper, our model inference is very efficient.

\paragraph{Training} Training is fast because we only turn on semantic augmentation and homography smoothness loss after 150k iterations (200k in total), similar to what have been done in SemARFlow~\cite{yuan2023semarflow}. The reasons are as follows.
\begin{itemize}
    \item Both semantic augmentation and homography smoothness loss rely on the current flow estimate to generate self-supervised loss signals, so we need to use flow at a later checkpoint to make sure they are reliable. Otherwise, the loss signals could be misleading.
    \item Semantic augmentation can generate very challenging self-supervised samples, which is better to be used at a later stage.
\end{itemize}

For the mask feature module adaptation, the added network size is very small (111.9k) since most of the added modules are 1-by-1 convolutions. Our typical full experiment training time is around 64 hours on 8 V100 GPUs.

We would like to emphasize that our goal is to investigate how SAM-style segmentations can benefit optical flow estimation. Optimizing SAM efficiency is outside the scope of this paper.
%% WARNING: end of supplementary pages 

% \clearpage
% \newpage
% {
%     \small
%     \bibliographystyle{format_cvpr2024/ieeenat_fullname}
%     \bibliography{refs}
% }

\end{document}